\documentclass[10pt,twocolumn,letterpaper]{article}

\usepackage[pagenumbers]{cvpr} 

\usepackage{multirow}

\definecolor{cvprblue}{rgb}{0.21,0.49,0.74}
\usepackage[pagebackref,breaklinks,colorlinks,allcolors=cvprblue]{hyperref}
\DeclareMathOperator*{\argmax}{argmax}

\def\paperID{10597} 
\def\confName{CVPR}
\def\confYear{2025}

\title{FEVER-OOD: Free Energy Vulnerability Elimination for Robust Out-of-Distribution Detection\vspace{-0.4cm}}

\author{Brian K.S. Isaac-Medina\\
Durham University\\
Durham, UK\\
{\tt\small brian.k.isaac-medina@durham.ac.uk}
\and
Mauricio Che\\
University of Vienna\\
Vienna, Austria\\
{\tt\small mauricio.adrian.che.moguel@univie.ac.at}
\and
Yona F.A. Gaus\\
Durham University\\
Durham, UK\\
{\tt\small yona.f.binti-abd-gaus@durham.ac.uk}
\and
Samet Akcay\\
Intel Corporation\\
Swindon, UK\\
{\tt\small samet.akcay@intel.com}
\and
Toby P. Brekcon\\
Durham University\\
Durham, UK\\
{\tt\small toby.breckon@durham.ac.uk}
}

\begin{document}
\maketitle
\begin{abstract}
Modern machine learning models, that excel on computer vision tasks such as classification and object detection, are often overconfident in their predictions for Out-of-Distribution (OOD) examples, resulting in unpredictable behaviour for open-set environments. Recent works have demonstrated that the free energy score is an effective measure of uncertainty for OOD detection given its close relationship to the data distribution. However, despite free energy-based methods representing a significant empirical advance in OOD detection, our theoretical analysis reveals previously unexplored and inherent vulnerabilities within the free energy score formulation such that in-distribution and OOD instances can have distinct feature representations yet identical free energy scores. This phenomenon occurs when the vector direction representing the feature space difference between the in-distribution and OOD sample lies within the null space of the last layer of a neural-based classifier. To mitigate these issues, we explore lower-dimensional feature spaces to reduce the null space footprint and introduce novel regularisation to maximize the least singular value of the final linear layer, hence enhancing inter-sample free energy separation. We refer to these techniques as \textit{Free Energy Vulnerability Elimination for Robust Out-of-Distribution Detection (FEVER-OOD)}. Our experiments show that FEVER-OOD techniques achieve state of the art OOD detection in Imagenet-100, with average OOD false positive rate (at 95\% true positive rate) of 35.83\% when used with the baseline Dream-OOD model.
\end{abstract}
\vspace{-1cm}    
\vspace{-0.1cm}
\section{Introduction}
\label{sec:intro}

\begin{figure}[t]
    \centering
    \includegraphics[width=\columnwidth]{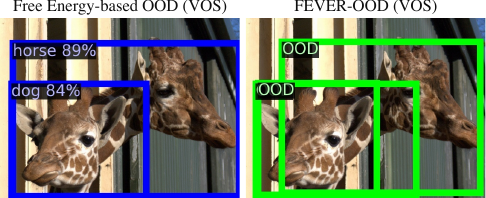}
    \vspace{-4mm}
    \caption{FEVER-OOD (right, \textcolor{green}{green}) improves baseline Free Energy-based OOD detection methods (left, \textcolor{blue}{blue}).}
    \label{fig:placeholder1}
    \vspace{-5mm}
\end{figure}

Out-of-distribution (OOD) detection (c.f. anomaly/outlier detection), aims to identify abnormal samples that deviate significantly from a given data distribution. Whilst OOD is essential in developing deployable machine learning systems, to ensure valid inference only on data drawn from same distribution used in training \cite{yang2024generalized, salehi2021unified, lang2023survey}, real-world deployments often involve scenarios where models encounter unseen OOD classes that could otherwise undermine decision stability and hence wider system robustness.

Methods such as deep ensembles \cite{lakshminarayanan2017simple}, ODIN \cite{liang2017enhancing}, Mahalanobis distance-based detection \cite{lee2018simple}, and generalized ODIN \cite{hsu2020generalized} all utilize \textit{softmax()} to differentiate in-distribution from OOD. Whilst intuitive, such approaches often struggle as deep neural networks commonly yield overly high \textit{softmax()} scores, even for inputs far from the training data distribution \cite{nguyen2015deep}. Given OOD and in-distribution samples should have significant spatial separation in any given feature space, Liu \etal \cite{liu2020energy} developed an alternative scoring function to effectively measure the alignment of a given sample to an \textit{a priori} learned distribution. Denoted as the free energy scoring function, 
this formulation demonstrates superior OOD performance across both classification \cite{duvos,du2023dream,du2024does} and detection tasks \cite{duvos,du2022unknown,kumar2023normalizing,isaac2024towards}.


Despite free energy-based methods representing a significant empirical advance in OOD detection, our theoretical analysis reveals previously unexplored  and inherent vulnerabilities within the free energy score formulation that could impact their practical reliability. These vulnerabilities reveal that in-distribution and OOD instances can have distinct feature representations yet produce identical free energy scores. This phenomenon occurs in the feature space when the vector direction representing the difference between an in-distribution and an OOD sample lies within the null space of the last layer of neural-based classifier. Such regions, effectively \textit{``blind spots"} in terms of OOD, lead to OOD and in-distribution generating identical or near-identical free energy score values.


To mitigate these issues, we explore lower-dimensional feature spaces to effectively reduce the size of the null space. In addition, we also introduce a novel regulariser to maximize the least singular value of the final linear layer, hence enhancing inter-sample free energy separation to improve OOD robustness. We refer to these techniques as \textit{Free energy Vulnerability Elimination for Robust Out-of-Distribution Detection (FEVER-OOD)}.

  

\noindent
In summary, our key contributions in this paper are:
\begin{itemize} \item The identification of inherent free energy score vulnerabilities, where under \textit{null-space} conditions, in-distribution and OOD instances will yield similar free energy scores despite differing feature representations. Denoted as \textbf{Null Space Vulnerabilities} (NSV) such conditions readily occur when the feature dimension exceeds the number of classes. Additionally, such free energy similarity can be influenced by the least singular vector (LSV) of the last linear layer of a neural-based classifier, leading to what we denote as \textbf{Least Singular Value Vulnerabilities} (LSVV). Both vulnerabilities negatively impact in-distribution to OOD discrimination.

\item Novel formulations that mitigate the NSV and LSVV impact (\cref{fig:placeholder1}). To address NSV, we introduce an additional layer to reduce the null space whilst for LSVV, we develop an \textbf{LSV Regulariser} that maximizes the least singular value, guaranteeing detectable energy variations from small feature-space differences and hence enhancing inter-sample energy scores discrimination. Furthermore, we explore a \textbf{Condition Number Regulariser} to promote balanced energy changes for displacements in all feature-space direction hence ensuring improved uniformity in energy score variations.

\item Comprehensive experiments and ablation studies using established benchmark datasets over both object classification and detection tasks, that demonstrate the effectiveness of FEVER-OOD in improving OOD detection performance. Specifically, applying FEVER-OOD to Dream-OOD \cite{du2023dream} for Imagenet-100 \cite{ILSVRC15} as in-distribution, we achieve a 10.13\% decrease in the average false positive rate (35.93\% vs. 39.98\%) and a 1.6\% increase in the average AUROC (93.12 vs 91.64), achieving state-of-the-art performance for OOD detection. \end{itemize}

\vspace{-0.2cm}
\section{Related Work}
\label{sec:related_work}
Identifying OOD samples is a crucial yet inherently challenging problem, as models are not exposed to, and hence cannot reliably differentiate, OOD samples.
\vspace{-0.2cm}
\paragraph{OOD Detection:} to overcome this challenge, researchers have begun incorporating additional data during training to establish a more conservative and safe decision boundary against OOD inputs. Some methods \cite{katz2022training, bai2023feed} have used unlabeled data to regularise model training while still focusing on classifying labeled in-distribution data, while others \cite{bai2024aha} leverage human assistance to strategically label examples within a novel maximum disambiguation region. Despite its potential, leveraging unlabeled data is non-trivial due to the heterogeneous mix of in-distribution and OOD instances. The absence of a clean OOD set presents significant challenges in designing effective OOD learning algorithms. A different branch of works incorporate post-hoc methods, focusing on designing OOD scoring mechanisms at test time. Hendrycks and Gimpel \cite{hendrycks2016baseline} introduced the maximum predicted \textit{softmax()} probability as an in-distribution score function, establishing an initial baseline for scoring OOD detection. Similarly, ODIN \cite{liang2017enhancing} aims to widen the gap between values in the \textit{softmax()} score vectors by temperature scaling and gradient-based input perturbation, enhancing the separation between in-distribution and OOD instances. Lee \etal \cite{lee2018simple} use the Mahalanobis distance between the average \textit{softmax()} vectors of images and incorporate intermediate features to improve detection performance and robustness. Nonetheless, the misalignment between the \textit{softmax()} score and the true data density \cite{liu2020energy, peng2024conjnorm} ultimately makes these techniques a suboptimal solution to OOD detection.
\vspace{-0.6cm}
\paragraph{Energy-based OOD Detection:} to address these limitations, Liu \etal \cite{liu2020energy} propose to use a free energy-based in-distribution score function instead of the \textit{softmax()} score to identify OOD instances. This approach regularises the model to produce lower energy values for in-distribution and higher energy values for auxiliary outliers, creating a significant energy gap that enhances the model ability to detect OOD samples. 
Energy-based OOD modeling has enabled novel outlier synthesis paradigms. For instance, Du \etal \cite{duvos} propose virtual outlier synthesis (VOS) by sampling outliers from low-likelihood in-distribution regions. VOS has shown to be effective for classification and object-level OOD detection. Further outlier synthesis techniques based on VOS have been proposed, including using normalizing flows \cite{kumar2023normalizing}, sampling based on the in-distribution boundary instances \cite{taonon} or extending VOS when class labels are unavailable \cite{isaac2024towards}. In addition to generating outliers in the feature space, Dream-OOD \cite{du2023dream} uses Stable Diffusion \cite{rombach2022high} to generate outliers in the pixel-space, learning better discriminative free-score values for OOD instances and providing visual examples of outliers. Whilst the core idea of energy score is that OOD must have  a significantly different free energy than in-distribution sample, we examine some inherent vulnerabilities in the energy formulation based on the null space and least singular value of the last linear layer of neural-based classifiers. 

\vspace{-0.2cm}
\section{Preliminaries: Free Energy Score}
\label{sec:preliminaries}

The goal of OOD detection is to detect outliers, $\mathbf{v}$, that have a low probability in the true (\textit{unknown}) in-distribution density $p(\mathbf{x})$. In this sense, OOD data is not available for training, meaning that OOD detection must be done based solely on $p(\mathbf{x})$. However, the computation of $p(\mathbf{x})$ is usually unfeasible or computationally untractable, so we instead resort to additional information. Now consider that we also have a set of in-distribution categories $\left\{1,\dots,K\right\}$ such that if we know the joint probability distribution, $p(\mathbf{x}, k)$, for $k\in \{1,\dots, K\}$, we could marginalize $p(\mathbf{x})$. Nonetheless, since the actual joint probabilities $p(\mathbf{x}, k)$ might be also untractable, we could instead use non-negative arbitrary \textit{potential functions} $\phi(\mathbf{x}, k)$ that measure the \textit{affinity} of the states of $\mathbf{x}$ and $k$ \cite{Goodfellow-et-al-2016}. In this sense, we can construct the unnormalized probability:
\begin{equation} \label{eq:unnormalized_prob}
    \Tilde{p}(\mathbf{x}) = \sum_{k=1}^{K} \phi(\mathbf{x}, k)
\end{equation}
that gives an idea of the likelihood $\mathbf{x}$.
Furthermore, energy-based models enforce the non-negative constraint of $\phi$ by:
\begin{equation} \label{eq:EBM}
\phi(\mathbf{x}, k)=\exp\left(-E(\mathbf{x}, k)\right)\,,    
\end{equation} 
where $E$ is an unbounded arbitrary function called the \textit{energy}. By substituting \cref{eq:EBM} into \cref{eq:unnormalized_prob} and taking the negative $\log$\footnote{N.B. the signs of the free energy and the argument of the exponent of $\phi$ are arbitrarily negative for historical reasons.}, we get a value known as the free energy (from the \textit{Helmholtz free energy} in thermodynamics):
\begin{equation} \label{eq:free-energy}
    \mathcal{F}(\mathbf{x}) = -\log \Tilde{p}(\mathbf{x}) = -\log \sum_{k=1}^{K} \exp(-E(\mathbf{x}, k))\,,
\end{equation}
and more recently used as a concept in OOD detection \cite{duvos,liu2020energy,kumar2023normalizing,isaac2024towards,du2023dream}. To get the actual density $p(\mathbf{x})$, we must compute the normalizing partition function of \cref{eq:unnormalized_prob}; since we are interested in distinguishing between in-distribution and OOD, $\mathcal{F}(\mathbf{x})$ thus serves well for this purpose. 

In a general machine learning context, a classifier $f({\mathbf{x}; \theta)}:\mathbb{R}^d\rightarrow\mathbb{R}^K$ with learned parameters $\theta$ assigns categorical unnormalized probabilities $f_k(\mathbf{x}; \theta)$, called \textit{logits}, for a given $\mathbf{x}\in\mathbb{R}^d$. While the logits are usually normalized using the \textit{softmax()} function in order to obtain a probability distribution, they can also be used as the energy $E$ in \cref{eq:EBM}. This realization enables the use of the free energy (\cref{eq:free-energy}) as a measure of uncertainty for a classifier:
\begin{equation} \label{eq:free-energy-ood}
    \mathcal{F}(\mathbf{x}) = -\log \sum_{k=1}^{K} \exp(f_k(\mathbf{x}; \theta))\,.
\end{equation}
Liu \etal \cite{liu2020energy} demonstrate that \cref{eq:free-energy-ood} is superior for effective OOD detection when compared to use of \textit{softmax()} whilst Du \etal \cite{duvos} explicitly use the free energy score (\cref{eq:free-energy-ood}) to perform binary classification between in-distribution and OOD by training with synthetic outliers. 


\section{Vulnerabilities of the Free Energy Score}
\begin{figure}[t]
    \centering
    \includegraphics[width=\columnwidth]{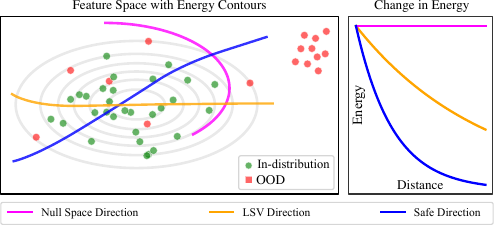}
    \caption{Vulnerabilities of Free Energy-based OOD detection. Directions in the \textcolor{magenta}{Null Space} do not change the energy, while the \textcolor{orange}{LSV direction} has minimal change, compared to  \textcolor{blue}{other directions}.}
    \label{fig:placeholder2}
    \vspace{-5mm}
\end{figure}
\label{sec:vulnerability}
Although the free energy (\cref{eq:free-energy-ood}) is an effective uncertainty measure for OOD detection \cite{liu2020energy, duvos}, here we examine some inherent vulnerabilities when used in conjunction with a neural network based classifier. Firstly, \cref{sec:ns_last_layer} investigates OOD instances that have similar energies as in-distribution samples, whilst \cref{sec:minimal_energy} discusses minimal changes in free energy, and finally \cref{sec:ns_reduction} outlines our proposed approach to minimize the impact of these vulnerabilities.

\subsection{Null Space Vulnerability}\label{sec:ns_last_layer}
Consider a neural network-based classifier $f=(g \circ h)(\mathbf{x})$, where $h(\mathbf{x}): \mathbb{R}^d \rightarrow \mathbb{R}^{d'}$ encodes the input into a $d'$-dimensional feature vector and $g: \mathbb{R}^{d'}\rightarrow \mathbb{R}^K$ is the last layer of the neural network that transforms the features into unnormalized class probabilities (\ie, the logits). If $g$ is a linear layer (as for most classifiers \cite{he2016deep,xie2020self,dosovitskiy2021an}), then:
\begin{equation}
    \label{eq:last-linear-layer}
    f(\mathbf{x})= \mathrm{W}^\top_\mathit{cls} h(\mathbf{x})\,,
\end{equation}
where $\mathrm{W}_\mathit{cls} \in \mathbb{R}^{d' \times K}$ are learnable weights. Hence, the logits $f_k(\mathbf{x};\theta)$ 
correspond to the $k$-th index of $f(\mathbf{x})$.

The core idea of energy-based OOD is that outliers $\mathbf{v}$ have a significantly different free energy, $\mathcal{F}(\mathbf{v})$, from in-distribution samples, $\mathcal{F}(\mathbf{x})$. Since OOD instances should be far from in-distribution samples in the feature space \cite{duvos,taonon}, the expected distance from $h(\mathbf{x})$ and $h(\mathbf{v})$ should be greater than a given boundary distance, $d_b$, as follows:
\begin{equation} \label{eq:boundary-distance}
    \mathbb{E}\left[\lVert h(\mathbf{x}) - h(\mathbf{v})\rVert\right] \geq d_b\,,
\end{equation}
for a norm $\lVert \cdot \rVert$. Subsequently, we can rewrite \cref{eq:boundary-distance} as:
\begin{equation} \label{eq:outlier-as-perturbation}
    h(\mathbf{v}) = h(\mathbf{x}) + \boldsymbol{\delta}\,,
\end{equation}
where $\boldsymbol{\delta}$ is a vector with $\mathbb{E}\left[\lVert \boldsymbol{\delta} \rVert\right] \geq d_b$. By substituting \cref{eq:outlier-as-perturbation} into \cref{eq:last-linear-layer} we obtain the following:
\begin{equation} \label{eq:outlier-last-layer}
        f(\mathbf{v}) = \mathrm{W}^\top_\mathit{cls} \left(h(\mathbf{x}) + \boldsymbol{\delta}\right) = f(\mathbf{x}) + \mathrm{W}^\top_\mathit{cls}\boldsymbol{\delta}\,.
\end{equation}
However within this formulation, if $\mathrm{W}_\mathit{cls}$ has more rows than columns, which occurs when the feature dimension is larger than the number of classes, the size of the null space of $\mathrm{W}^\top_\mathit{cls}$, called the \textit{nullity}, is $\mathrm{nullity}(\mathrm{W}^\top_\mathit{cls})=d' - \mathrm{rank}(\mathrm{W}_\mathit{cls})$ due to the rank-nullity theorem. 
Mathematically, this means that there exists a linear subspace $\mathrm{Null}(\mathrm{W}^\top_\mathit{cls}) \subset \mathbb{R}^{d'}$ such that if $\boldsymbol{\delta} \in \mathrm{Null}(\mathrm{W}^\top_\mathit{cls})$ then $\mathrm{W}^\top_\mathit{cls} \boldsymbol{\delta}=\boldsymbol{0}$. Within our OOD context, if an outlier instance, $\mathbf{v}$, has a feature representation, $h(\mathbf{v})$, such that its difference with respect to the inlier feature representation, $h(\mathbf{x})$,  of \textit{any} inlier instance, $\mathbf{x}$, is in the null space of the last layer of the classifier, then their free energy score are the same:
\begin{equation} \label{eq:null-space-vulnerability}
h(\mathbf{v}) - h(\mathbf{x}) \in \mathrm{Null}(\mathrm{W}^\top_\mathit{cls}) \Rightarrow \mathcal{F}(\mathbf{x}) = \mathcal{F}(\mathbf{v})\,.
\end{equation}
Practically, \cref{eq:null-space-vulnerability} is telling us that it is possible for in-distribution and OOD instances to have different feature representations (satisfying the condition of \cref{eq:boundary-distance}) and yet have the same free energy score, as depicted in \cref{fig:placeholder2}. This fundamentally challenges the notion that free energy is an effective mechanism for OOD detection as any approaches reliant on \cref{eq:free-energy}, such as  \cite{liu2020energy, duvos, kumar2023normalizing, du2023dream,isaac2024towards}, contain inherent \textit{``detection blind-spots''} as per \cref{eq:null-space-vulnerability}. We denote \cref{eq:null-space-vulnerability} as the \textbf{Null Space Vulnerabilities} (NSV) of classifier $f$ with size $\mathrm{nullity}(\mathrm{W}^\top_\mathit{cls})= d' - \mathrm{rank}(\mathrm{W}_\mathit{cls})$. An approach to minimize this vulnerability is outlined in \cref{sec:ns_reduction}.

\subsection{Minimal Change of Free Energy}\label{sec:minimal_energy}
We now turn our attention to the case where $\boldsymbol{\delta}$ is not in the null space of $\mathrm{W}^\top_\mathit{cls}$. Furthermore, we can focus on $\boldsymbol{\delta}$ orthogonal to $\mathrm{Null}(\mathrm{W}^\top_\mathit{cls})$, since every vector in the domain of $\mathrm{W}^\top_\mathit{cls}$ can be decomposed into one component in $\mathrm{Null}(\mathrm{W}^\top_\mathit{cls})$ and one component in $\mathrm{Null}(\mathrm{W}^\top_\mathit{cls})^\perp$, and the effect of the component in $\mathrm{Null}(\mathrm{W}^\top_\mathit{cls})$ will be dealt with using the null space reduction method (\cref{sec:ns_reduction}).

In order to create more discriminative OOD detectors based on the free energy score, we investigate cases where the difference in OOD and in-distribution free energies is minimal. We can state our problem formulation to find the case that minimizes the free energy difference:
\begin{equation}
    \label{eq:min-change-energy}
    \begin{gathered}
    \min_{\substack{\boldsymbol{\delta}: \lVert \boldsymbol{\delta} \rVert \ge d_b,\\\boldsymbol{\delta} \in \mathrm{Null}(\mathrm{W}^\top_\mathit{cls})^\perp}} \; \lvert \mathcal{F}(\mathbf{v}) - \mathcal{F}(\mathbf{x}) \rvert 
    \\
    = \min_{\substack{\boldsymbol{\delta}: \lVert \boldsymbol{\delta} \rVert \ge d_b,\\\boldsymbol{\delta} \in \mathrm{Null}(\mathrm{W}^\top_\mathit{cls})^\perp}} \; \left\lvert \log \frac{\sum_k^K \exp(f_k(h(\mathbf{x})))}{\sum_k^K \exp(f_k(h(\mathbf{x}) + \boldsymbol{\delta}))} \right\rvert\,.
    \end{gathered}
\end{equation}

Whilst in \cref{eq:min-change-energy} we set $\lVert \boldsymbol{\delta} \rVert \ge d_b$ instead of $\mathbb{E}\left[\lVert \boldsymbol{\delta} \rVert\right] \geq d_b$, in reality we are just interested in the case of non-zero $\boldsymbol \delta$ since otherwise the solution would be trivial. Whilst the solution to \cref{eq:min-change-energy} may be challenging to obtain, we note the difference in free-energies tends to 0 if the value inside the $\log()$ function tends to 1. This is equivalent to making $f_k(h(\mathbf{x}))$ and $f_k(h(\mathbf{x}) + \boldsymbol{\delta})$ similar for all $k$. Therefore, considering that $f_k$ is in the form of \cref{eq:last-linear-layer}, 
our problem consists on finding the following proxy minimum:
\begin{equation} \label{eq:min-last-lin-layer}
    \begin{gathered}
    \min_{\substack{\boldsymbol{\delta}: \lVert \boldsymbol{\delta} \rVert \ge d_b,\\\boldsymbol{\delta} \in \mathrm{Null}(\mathrm{W}^\top_\mathit{cls})^\perp}} \lVert \mathrm{W}^\top_\mathit{cls}(h(\mathbf{x}) + \boldsymbol{\delta}) - \mathrm{W}^\top_\mathit{cls}h(\mathbf{x}) \rVert \\ = \min_{\substack{\boldsymbol{\delta}: \lVert \boldsymbol{\delta} \rVert \ge d_b,\\\boldsymbol{\delta} \in \mathrm{Null}(\mathrm{W}^\top_\mathit{cls})^\perp}} \lVert \mathrm{W}^\top_\mathit{cls}\boldsymbol{\delta}\rVert\,.
    \end{gathered}
\end{equation}
For the $L^2$-norm in \cref{eq:min-last-lin-layer}, we thus obtain:
\begin{equation}
    \label{eq:smin}
    \min_{\substack{\boldsymbol{\delta}: \lVert \boldsymbol{\delta} \rVert_2 \ge d_b,\\\boldsymbol{\delta} \in \mathrm{Null}(\mathrm{W}^\top_\mathit{cls})^\perp}} \lVert \mathrm{W}^\top_\mathit{cls}\boldsymbol{\delta}\rVert_2 = d_b \sigma_\mathit{min}(\mathrm{W}_\mathit{cls})\,,
\end{equation}
where $\sigma_\mathit{min}(\mathrm{W}_\mathit{cls})$ is the least singular value of $\mathrm{W}_\mathit{cls}$ (see further mathematical formulation details in Supp. Mat.). 

Subsequently, \cref{eq:smin} holds for the case where $\boldsymbol{\delta}$ is in the direction corresponding to the least singular value, such that the change in the free energy will be smaller than any other direction for $\boldsymbol{\delta}$ (and considering $\boldsymbol{\delta} \in \mathrm{Null}(\mathrm{W}^\top_\mathit{cls})^\perp$). 
We name this result (from  \cref{eq:smin}) the \textbf{Least Singular Value Vulnerabilities} (LSVV) of $f$. In our OOD context, it means that there could exist some outliers far from the inlier distribution, in terms of \cref{eq:boundary-distance}, with similar free energy scores because they may fall in the direction of the least singular value of $\mathrm{W}_\mathit{cls}$. Again, this fundamentally challenges the notion that free energy is an effective mechanism for OOD as free energy has been shown to be non-linear with respect to a given feature representation, $h()$.

This analysis is itself inspired by the seminal adversarial attacks analysis of Simon-Gabriel \etal \cite{pmlr-v97-simon-gabriel19a}, where they instead maximize the change in the classifier response for small input perturbations. Whilst here we are modelling the opposite problem, (\ie, the minimum change in the classifier for perturbations larger than a non-zero value), we could consider this analysis as an effective adversarial attack mechanism for the OOD domain. 

\subsection{Null Space Reduction \& Regularisation}\label{sec:ns_reduction}
In this section, we introduce our Free Energy Vulnerability Elimination for Robust OOD Detection (FEVER-OOD) framework, where we explore readily available methods to minimize the NSV (\cref{sec:ns_last_layer}) and LSVV (\cref{sec:minimal_energy}) impact potential within free energy score based OOD approaches.

\begin{figure}[t]
    \centering
    \includegraphics[width=0.95\columnwidth]{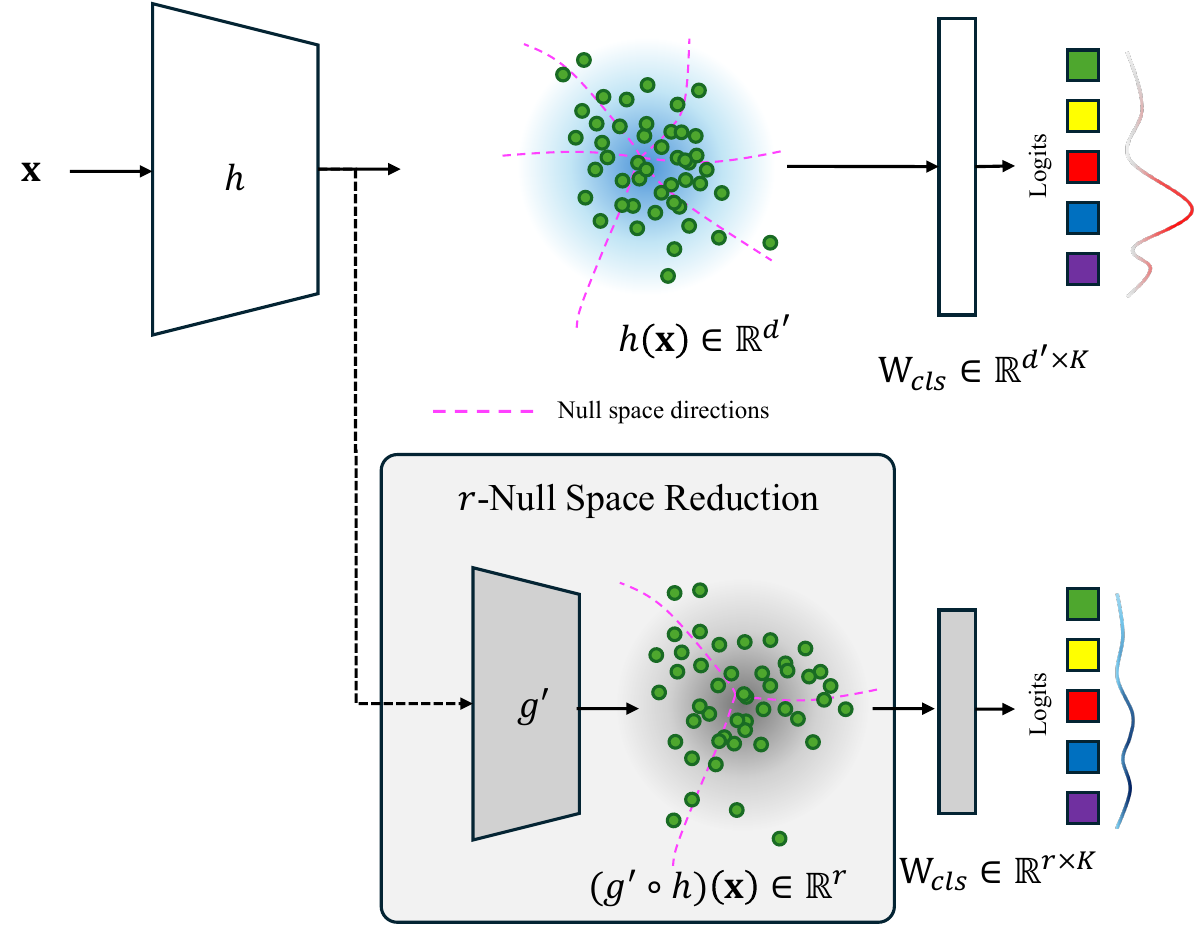}
    \vspace{-2mm}
    \caption{Null space reduction. We add an extra layer $g'$ to create $r$-dimensional features, reducing the size of the null space.}
    \label{fig:placeholder3}
    \vspace{-5mm}
\end{figure}

\noindent
\textbf{Null Space Reduction (NSR)}. For NSV, we propose to reduce the null space by adding an extra linear layer $g': \mathbb{R}^{d'} \rightarrow \mathbb{R}^r$, where we choose $r < d'$ to reduce the size of the NSV (\cref{fig:placeholder3}). We denote this method as $r$-null space reduction ($r$-NSR). Therefore, the new classifier becomes $f'=g \circ g' \circ h$ such that the feature representation of a sample is in $\mathbb{R}^{r}$ and is given instead by $(g' \circ h)(\mathbf{x})$. Effectively, within an OOD context, we are promoting better discrimination between OOD and in-distribution instances. 

\noindent
\textbf{Least Singular Value Regulariser (LSVR)}. To minimize the impact of LSVV, we maximise $\sigma_\mathit{min}(\mathrm{W}_\mathit{cls})$ by adding an LSVR to the loss function $\mathcal{L}$ of a baseline method:
\begin{equation} \label{eq:lsvv-reg}
    \mathcal{L}_\mathit{LSV} := \mathcal{L} + \lambda_\mathit{LSV} \sigma^{-1}_\mathit{min}(\mathrm{W}_\mathit{cls})\,,
\end{equation}
where $\lambda_\mathit{LSV}$ is a scalar hyperparameter that controls the contribution of the LSV regulariser. It is additionally noted that alternatively maximizing the LSV might also increase the rest of the singular values, resulting in more significant changes of energy in all directions. In the context of OOD detection, LSVR increases the discriminative capacity for OOD detection of baseline methods. 

\noindent
\textbf{Condition Number (CN) Regulariser}. Whilst the LSV regulariser will increase the change in energy for $\boldsymbol{\delta}$ in the LSV direction, there is no constraint with regard to the distribution over all other directions. For this reason, we introduce the Condition Number Regulariser (CNR):
\begin{equation} \label{eq:cn-reg}
    \mathcal{L}_\mathit{CN} := \mathcal{L} + \lambda_\mathit{CN} \kappa(\mathrm{W}_\mathit{cls})\,,
\end{equation}
where $\kappa(\mathrm{W}_\mathit{cls})$ is the condition number of the $\mathrm{W}_\mathit{cls}$ matrix, \ie, the ratio of the greatest and least singular values. Intuitively, this will encourage equalized energy changes for displacements in all directions where we note that \cref{eq:cn-reg} similarly assumes use of the $L^2$-norm as per \cref{eq:smin}.


\vspace{-0.2cm}
\section{Experimental Details}
\label{sec:experiments}

We study the impact of FEVER-OOD on free energy baseline methods (\cref{sec:exp-baseline-methods}) among different standard datasets (\cref{sec:exp-dataset}). Implementation details are given in \cref{sec:exp-implementation}.

\subsection{Baseline Methods and Evaluation Metrics} \label{sec:exp-baseline-methods}
Three free energy baseline methods for image OOD detection that use the free energy score are tested: VOS \cite{duvos}, FFS \cite{kumar2023normalizing} and Dream-OOD \cite{du2023dream}. These baseline methods learn a small MLP $\psi:\mathbb{R}\rightarrow \left(0, 1\right)$ that uses the free energy in \cref{eq:free-energy-ood} to predict an uncertainty score. Since OOD data is unavailable during training, these methods use synthetic outliers to train $\psi$. While VOS and FFS synthesize outliers in the feature space $h(\mathbf{x})$, Dream-OOD generates outliers in the pixel space using Stable Diffusion \cite{rombach2022high}. Additionally, VOS and FFS are also implemented as object detectors for object-level OOD detection, synthesizing outliers before the last layer of the classification head of a Faster RCNN \cite{faster-rcnn}.

\begin{table*}[ht]
\centering
\caption{CIFAR-10 Results (ID Acc. $=$ in-distribution accuracy; Null Space Reduction (NSR) methods $=$ our approach).}
\vspace{-0.2cm}
\label{table:cifar-10-eval}
\resizebox{0.95\linewidth}{!}{%
\renewcommand{\arraystretch}{1}
\begin{tabular}{lcccccccccccccccl}
\hline
& \multicolumn{3}{c}{\multirow{2}{*}{FEVER-OOD}} &\multicolumn{12}{c}{OOD Datasets}&\\ 
\cline{5-16}

 &&&& \multicolumn{2}{c}{Textures}& \multicolumn{2}{c}{SHVN}& \multicolumn{2}{c}{Places365}& \multicolumn{2}{c}{LSUN}& \multicolumn{2}{c}{iSUN}& \multicolumn{2}{c}{Avg}& \\ 
 
 \cline{5-16}
 
\multirow{-3}{*}{Method}         & $r$-NSR & $\lambda_\mathit{LSV}$ & $\lambda_\mathit{CN}$ & FPR95 & AUROC & FPR95 & AUROC & FPR95 & AUROC & FPR95 & AUROC & FPR95 & AUROC & FPR95 & AUROC &           \multirow{-3}{*}{ID Acc}               \\

\hline

\rowcolor[HTML]{EFEFEF} 
VOS      &-&-&-& 50.62& 86.98& 42.47& 90.53& 40.17& 89.72& 7.94& 98.34& 25.30& 95.28& 33.30& 92.17                     & 94.85                    \\
\rowcolor[HTML]{EFEFEF} 
&     -  & 1.0                                       &     -                                    & 48.12                     & 88.43                     & 39.59                     & 91.64                     & 34.42                     & 92.15                     & 6.51                      & 98.49                     & 17.14                     & 96.84                     & 29.16                     & 93.51                     & 94.48                    \\
\rowcolor[HTML]{EFEFEF} 
&    -   &   -                                        & 0.1                                     & 47.01                     & 86.15                     & 29.54                     & 93.61                     & 42.00                     & 88.45                     & 7.71                      & 98.31                     & 29.82                     & 94.47                     & 31.22                     & 92.20                     & 94.63                    \\

&     96                  &      -                                     &         -                                & 50.59                     & 88.79                     & \textbf{25.49}            & \textbf{95.26}            & 39.09                     & 91.90                     & 7.36                      & 98.42                     & 26.90                     & 95.55                     & 29.89                     & 93.98                     & 94.69                    \\
&      96                   & 1.0                                       &     -                                    & \textbf{35.21}            & \textbf{93.40}            & 42.84                     & 93.12                     & \textbf{32.66}            & \textbf{93.37}            & \textbf{3.24}             & \textbf{99.26}            & \textbf{13.44}            & \textbf{97.65}            & \textbf{25.48}            & \textbf{95.36}            & 94.84                    \\
&      96               &           -                                & 0.1                                     & 39.97                     & 91.77                     & 39.80                     & 93.96                     & 39.16                     & 90.69                     & 4.35                      & 99.00                     & 24.23                     & 95.59                     & 29.50                     & 94.20                     & 94.55                    \\
\rowcolor[HTML]{EFEFEF} 
&64 &                      -                     &               -                          & 45.05                     & 89.40                     & 45.37                     & 91.58                     & 36.74                     & 91.53                     & 6.36                      & 98.74                     & 22.48                     & 95.95                     & 31.20                     & 93.44                     & 94.73                    \\
\rowcolor[HTML]{EFEFEF} 
&64 & 1.0                                       &         -                                & 44.04                     & 90.53                     & 34.13                     & 95.16                     & 36.39                     & 92.43                     & 5.28                      & 98.74                     & 19.74                     & 96.66                     & 27.92                     & 94.70                     & 94.80                    \\
\rowcolor[HTML]{EFEFEF} 
&64 &                -                           & 0.001                                   & 42.12                     & 91.01                     & 43.18                     & 93.20                     & 36.23                     & 91.71                     & 6.66                      & 98.71                     & 25.40                     & 95.01                     & 30.72                     & 93.93                     & 94.75                    \\
& 32                        &             -                              &       -                                  & 38.39                     & 91.87                     & 66.94                     & 88.35                     & 37.35                     & 91.68                     & 5.73                      & 98.86                     & 29.84                     & 94.58                     & 35.65                     & 93.07                     & 94.64                    \\
& 32                        & 0.001                                     &       -                                  & 47.80                     & 89.19                     & 40.57                     & 93.90                     & 38.48                     & 91.78                     & 5.95                      & 98.73                     & 23.04                     & 95.94                     & 31.17                     & 93.91                     & 94.78                    \\
& 32                        &          -                                 & 0.1                                     & 43.86                     & 89.65                     & 28.08                     & 95.23                     & 41.33                     & 90.33                     & 5.31                      & 98.94                     & 29.42                     & 94.76                     & 29.60                     & 93.78                     & 95.05                    \\
\rowcolor[HTML]{EFEFEF} 
& 10&                         -                  &          -                               & 54.56                     & 88.86                     & 33.75                     & 94.80                     & 44.43                     & 90.98                     & 10.48                     & 98.25                     & 42.68                     & 92.81                     & 37.18                     & 93.14                     & 89.64                    \\
\rowcolor[HTML]{EFEFEF} 
&10 & 0.01                                      &        -                                 & 59.37                     & 86.68                     & 35.98                     & 94.36                     & 41.52                     & 89.65                     & 10.48                     & 98.01                     & 32.60                     & 94.25                     & 35.99                     & 92.59                     & 94.34                    \\
\rowcolor[HTML]{EFEFEF} 
& 10&                   -                        & 0.001                                   & 59.02                     & 86.46                     & 67.06                     & 88.05                     & 52.25                     & 88.07                     & 19.99                     & 96.56                     & 28.31                     & 95.14                     & 45.33                     & 90.86                     & 88.73                    \\ \hline
FFS  &-                            &          -                                 &    -                                     & 46.01                     & 87.28                     & 48.73                     & 88.09                     & 41.39                     & 89.01                     & 5.96                      & 98.79                     & 27.71                     & 94.67                     & 33.96                     & 91.57                     & 94.65                    \\
&-& 0.001                                     &   -                                      & 43.94                     & 87.39                     & 22.05                     & 94.74                     & 40.12                     & 89.11                     & 3.21                      & 99.29                     & 16.88                     & 96.52                     & 25.24                     & 93.41                     & 95.08                    \\
&-&            -                               & 1.0                                     & 44.37                     & 88.12                     & 32.82                     & 92.39                     & 43.39                     & 88.35                     & 4.98                      & 98.92                     & \textbf{9.18}             & \textbf{98.11}            & 26.95                     & 93.18                     & 94.74                    \\
\rowcolor[HTML]{EFEFEF} 
 &96&        -                                   &     -                                    & 47.63                     & 88.64                     & 29.13                     & 95.29                     & 38.13                     & 91.64                     & 4.50                      & 99.03                     & 20.40                     & 96.50                     & 27.96                     & 94.22                     & 94.54                    \\
\rowcolor[HTML]{EFEFEF} 
 &96& 1.0                                       &           -                              & 43.00                     & \textbf{91.85}            & 30.14                     & 94.81                     & \textbf{32.81}            & \textbf{93.10}            & 4.09                      & 99.12                     & 14.61                     & 97.43                     & 24.93                     & \textbf{95.26}            & 94.81                    \\
\rowcolor[HTML]{EFEFEF} 
&96 &   -                                        & 0.001                                   & 44.37                     & 90.33                     & 28.56                     & 95.37                     & 39.05                     & 91.59                     & 4.59                      & 99.10                     & 24.73                     & 95.47                     & 28.26                     & 94.37                     & 94.83                    \\
&64&                -                           &           -                              & 48.24                     & 89.67                     & 45.38                     & 89.67                     & 38.43                     & 90.80                     & 3.99                      & 99.15                     & 33.55                     & 92.91                     & 33.92                     & 92.44                     & 94.77                    \\
&64& 1.0                                       &            -                             & 43.41                     & 90.96                     & 31.30                     & 93.96                     & 36.52                     & 92.07                     & 3.18                      & 99.22                     & 10.23                     & 98.00                     & 24.93                     & 94.84                     & 94.61                    \\
&64&                    -                       & 0.001                                   & 44.01                     & 90.36                     & \textbf{16.97}            & \textbf{96.71}            & 42.08                     & 89.65                     & 3.93                      & 99.20                     & 13.80                     & 97.43                     & \textbf{24.16}            & 94.67                     & 94.91                    \\
\rowcolor[HTML]{EFEFEF} 
&32&  -                                         &   -                                      & 40.88                     & 90.57                     & 38.84                     & 91.25                     & 43.32                     & 89.74                     & 4.13                      & 99.11                     & 26.67                     & 94.77                     & 30.77                     & 93.09                     & 94.57                    \\
\rowcolor[HTML]{EFEFEF} 
 &32& 0.001                                     &    -                                     & 45.18                     & 89.44                     & 28.32                     & 95.35                     & 38.69                     & 91.27                     & \textbf{3.01}             & \textbf{99.30}            & 24.83                     & 95.59                     & 28.01                     & 94.19                     & 94.69                    \\
\rowcolor[HTML]{EFEFEF} 
 &32&        -                                   & 0.01                                    & \textbf{42.26}            & 90.58                     & 33.15                     & 95.00                     & 42.16                     & 90.37                     & 3.91                      & 99.22                     & 21.71                     & 96.36                     & 28.64                     & 94.31                     & 94.82                    \\
&10&      -                                     &     -                                    & 57.67                     & 82.43                     & 46.68                     & 89.41                     & 46.81                     & 87.28                     & 12.56                     & 97.52                     & 28.66                     & 94.62                     & 38.48                     & 90.25                     & 94.72                    \\
&10& 0.001                                     &    -                                     & 56.40                     & 86.41                     & 46.42                     & 92.16                     & 39.34                     & 90.55                     & 6.71                      & 98.62                     & 22.77                     & 96.14                     & 34.33                     & 92.78                     & 94.55                    \\
&10&                          -                 & 0.001                                   & 55.59                     & 84.27                     & 74.97                     & 80.77                     & 36.36                     & 91.23                     & 9.75                      & 97.99                     & 31.69                     & 93.16                     & 41.67                     & 89.48                     & 94.49                    \\ \hline
\end{tabular}%
}
\renewcommand{\arraystretch}{1}
\vspace{-4mm}
\end{table*}

We follow standard evaluation metrics, reporting the false positive rate for OOD detection at 95\% true positive rate of in-distribution detection (FPR95) and the area under the receiver operating characteristic curve (AUROC). We also assess the effect of our proposed techniques into the main in-distribution task (accuracy for classification and MS-COCO \cite{coco_dataset} mean average precision (mAP) for object detection). FPR95 and AUROC metrics are reported using a single intersection-over-union threshold for object detection, corresponding to the one that maximises the F-1 score, as described by Harakeh and Waslander \cite{harakeh2021estimating}.

\subsection{Datasets} \label{sec:exp-dataset}
We assess FEVER-OOD with standard datasets for both classification and object detection. For image-level OOD detection (\ie, image classification), we train VOS and FFS using the CIFAR-10 and CIFAR-100 \cite{krizhevsky2009learning} datasets as in-distribution while testing OOD detection in five non-overlapping datasets: Textures \cite{textures_dataset}, SVHN \cite{svhn_dataset}, Places365 \cite{places365_dataset}, LSUN \cite{lsun_dataset} and iSUN \cite{isun_dataset}. We also test Dream-OOD with the CIFAR-100 dataset as in-distribution data with the same OOD datasets. Additionally, we also train Dream-OOD with the Imagenet-100 dataset, a 100 classes partition of Imagenet \cite{ILSVRC15}, as in-distribution while testing OOD detection in four datasets: Textures \cite{textures_dataset}, Places365 \cite{places365_dataset}, iNaturalist \cite{inaturalist} and SUN \cite{sun_dataset}. We use the PASCAL VOC \cite{voc_dataset} dataset as in-distribution for object-level OOD detection using VOS and FFS and test for OOD detection on the MS-COCO dataset \cite{coco_dataset} removing images containing objects with overlapping in-distribution categories \cite{duvos}.

\subsection{Implementation details} \label{sec:exp-implementation}
For image-level OOD detection with VOS and FFS, we use a WideResNet-40 \cite{zerhouni2017wide} with a 128-dimensional feature space. For CIFAR-10, we evaluate the $\{96,64,32,10\}$-NSR, while for CIFAR-100 we assess $\{114,100\}$-NSR. With respect to Dream-OOD models, we use a ResNet-34 \cite{he2016deep} with 512-dimensional feature space, evaluating $\{256,128,100\}$-NSR for both CIFAR-100 and Imagenet-100. Similarly, we reduce the feature space of the classification head to 768, 512 and 265 dimensions for VOS and FFS object detection experiments, having an original 1,024-dimensional object-wise feature space (as per Faster RCNN \cite{faster-rcnn}). With regards to the LSVR and CNR, we test with various loss weight values in all the architectures (1, 0.1, 0.01, and 0.001), to evaluate their impact on model performance (except for Imagenet-100, where we use $\lambda_\mathit{LSV}=0.001$ and $\lambda_\mathit{CN}=0.01$ based on CIFAR-100 experiments). We train all the models using the original corresponding baseline settings. Details on the hyperparameters and the training regime for each model are given in the Supp. Mat. To obtain reliable comparisons we use deterministic learning via a fixed random seed for all experiments. 
\vspace{-1.5mm}
\section{Results}
\vspace{-1.5mm}
\label{sec:results}
\begin{table*}[t]
\centering
\caption{CIFAR-100 Results (ID Acc. $=$ in-distribution accuracy; Null Space Reduction (NSR) methods $=$ our approach).}
\vspace{-3mm}
\label{table:cifar-100-eval}
\resizebox{0.95\linewidth}{!}{%
\renewcommand{\arraystretch}{1}
\begin{tabular}{lcccccccccccccccc}
\hline
& \multicolumn{3}{c}{\multirow{2}{*}{FEVER-OOD}} &\multicolumn{12}{c}{OOD Datasets}&\\
\cline{5-16}

 &&&& \multicolumn{2}{c}{Textures}& \multicolumn{2}{c}{SHVN}& \multicolumn{2}{c}{Places365}& \multicolumn{2}{c}{LSUN}& \multicolumn{2}{c}{iSUN}& \multicolumn{2}{c}{Avg}&  \\

 \cline{5-16}

\multirow{-3}{*}{Method}         & $r$-NSR & $\lambda_\mathit{LSV}$ & $\lambda_\mathit{CN}$ & FPR95 & AUROC & FPR95 & AUROC & FPR95 & AUROC & FPR95 & AUROC & FPR95 & AUROC & FPR95 & AUROC &   \multirow{-3}{*}{ID Acc}                       \\

\hline
\rowcolor[HTML]{EFEFEF}
VOS &   -        &     -                                      &     -                                    & 81.72                     & 75.83                     & 81.03                     & 83.81                     & 81.03                     & 76.29                     & 38.81                     & 92.94                     & 79.61                     & 75.04                     & 72.44                     & 80.78                     & 76.80                                             \\
\rowcolor[HTML]{EFEFEF}
&-              & 0.1                                  &     -                                   & \textbf{77.33}            & 78.22                     & 82.58                     & 84.06                     & \textbf{78.16}            & \textbf{77.86}            & 35.17                     & 93.35                     & \textbf{65.33}            & 82.69                     & 67.71                     & 83.24                     & 76.20 \\
\rowcolor[HTML]{EFEFEF}
&        -      &       -                                    & 0.01                                    & 82.14                     & 75.52                     & 72.76                     & 84.20                     & 81.20                     & 76.05                     & 39.90                     & 92.60                     & 84.14                     & 74.81                     & 72.03                     & 80.64                     & 76.15                                             \\
&         114                      &   -                                        &                -                         & 80.23                     & \textbf{79.12}            & 90.14                     & 76.47                     & 79.47                     & 77.17                     & 27.83                     & 95.04                     & 85.11                     & 74.45                     & 72.56                     & 80.45                     & 75.02                                             \\
&               114                & 0.01                                      &             -                            & 83.17                     & 77.05                     & \textbf{59.22}            & \textbf{89.61}            & 79.39                     & 76.35                     & 26.71                     & 95.49                     & 69.06                     & \textbf{84.99}            & \textbf{63.51}            & \textbf{84.70}            & 76.65                                             \\
&          114                     &     -                                      & 0.001                                   & 83.30                     & 77.78                     & 83.67                     & 80.59                     & 79.97                     & 77.82                     & 29.16                     & 95.01                     & 76.15                     & 80.30                     & 70.45                     & 82.30                     & 75.78                                             \\
\rowcolor[HTML]{EFEFEF}
&   100    &       -                                    &    -                                     & 83.71                     & 75.44                     & 88.12                     & 79.93                     & 80.63                     & 77.09                     & 30.97                     & 94.99                     & 81.65                     & 76.95                     & 73.02                     & 80.88                     & 75.90                                             \\
\rowcolor[HTML]{EFEFEF}
&   100    & 0.001                                     &     -                                    & 77.40                     & 78.43                     & 83.23                     & 77.67                     & \textbf{78.16}            & 76.46                     & \textbf{25.77}            & \textbf{95.60}            & 76.28                     & 80.35                     & 68.17                     & 81.70                     & 75.46                                             \\
\rowcolor[HTML]{EFEFEF}
&   100    &                                 -          & 1                                       & 100.00                    & 50.00                     & 100.00                    & 50.00                     & 100.00                    & 50.00                     & 100.00                    & 50.00                     & 100.00                    & 50.00                     & 100.00                    & 50.00                     & 1.00                                                 \\ \hline
FFS &           -                           &     -                                      & -                                        & 82.39                     & 75.11                     & 81.81                     & 81.60                     & 81.58                     & 75.86                     & 35.50                     & 93.42                     & 81.66                     & 73.01                     & 72.59                     & 79.80                     & 76.05                                             \\
&              -                        & 0.01                                      &      -                                   & 79.15                     & 78.53                     & 78.44                     & 82.65                     & 79.16                     & 77.85                     & 28.77                     & 95.26                     & \textbf{64.91}            & \textbf{84.01}            & 66.09                     & 83.66                     & 76.71                                             \\
&              -                        &                -                           & 0.001                                   & 78.75                     & 77.24                     & 72.36                     & 87.33                     & 81.24                     & 76.09                     & 34.62                     & 93.90                     & 75.34                     & 80.24                     & 68.46                     & 82.96                     & 76.45                                             \\
\rowcolor[HTML]{EFEFEF}
&    114   &               -                            &      -                                   & 80.61                     & 78.06                     & 90.25                     & 74.95                     & 81.49                     & 76.66                     & \textbf{22.74}            & \textbf{96.15}            & 81.33                     & 78.83                     & 71.28                     & 80.93                     & 75.33                                             \\
\rowcolor[HTML]{EFEFEF}
&    114   & 0.001                                     &        -                                 & \textbf{77.63}            & \textbf{79.29}            & \textbf{72.08}            & \textbf{87.41}            & \textbf{78.69}            & \textbf{78.12}            & 24.11                     & 95.87                     & 75.47                     & 80.54                     & \textbf{65.60}            & \textbf{84.25}            & 75.30                                             \\
\rowcolor[HTML]{EFEFEF}
&    114   &                       -                    & 0.001                                   & 82.06                     & 79.04                     & 82.55                     & 78.68                     & 80.18                     & 77.62                     & 27.45                     & 95.22                     & 86.15                     & 74.60                     & 71.68                     & 81.03                     & 74.89                                             \\
&        100                       &     -                                      &  -                                       & 81.20                     & 77.37                     & 84.27                     & 82.35                     & 80.87                     & 76.30                     & 23.83                     & 95.95                     & 80.53                     & 74.23                     & 70.14                     & 81.24                     & 76.03                                             \\
&        100                       & 0.001                                     &    -                                     & 80.82                     & 74.66                     & 83.54                     & 81.87                     & 81.01                     & 76.17                     & 24.96                     & 95.86                     & 78.11                     & 79.36                     & 69.69                     & 81.58                     & 75.54                                             \\
&        100                       &     -                                      & 1                                       & 100.00                    & 50.00                     & 100.00                    & 50.00                     & 100.00                    & 50.00                     & 100.00                    & 50.00                     & 100.00                    & 50.00                     & 100.00                    & 50.00                     & 1.00                                                 \\ \hline
\rowcolor[HTML]{EFEFEF}
Dream-OOD &     -   &             -                              &         -                                & 62.20                     & 83.84                     & 73.05                     & 84.56                     & 77.95                     & 79.43                     & 39.90                     & 92.87                     & 1.70                      & 99.58                     & 50.96                     & 88.06                     & 75.61 \\
\rowcolor[HTML]{EFEFEF}
&     -   & 0.01                                      &       -                                  & 58.45                     & 86.04                     & 68.75                     & 87.65                     & 77.45                     & 78.59                     & \textbf{15.45}            & \textbf{97.24}            & 1.55                      & 99.63                     & 44.33                     & 89.83                     & 75.87 \\
\rowcolor[HTML]{EFEFEF}
&     -   &      -                                     & 0.001                                   & 57.4                      & 86.28                     & 77.75                     & 85.13                     & 78.6                      & 78.73                     & 27.2                      & 95.01                     & 1.55                      & 99.57                     & 48.5                      & 88.94                     & 76.32 \\
& 256                        &     -                                      &       -                                  & 60.00                     & 85.42                     & 67.50                     & 85.84                     & 75.90                     & 79.57                     & 19.85                     & 96.74                     & \textbf{1.00}             & \textbf{99.78}            & 44.85                     & 89.47                     & 77.01                         \\
& 256                        &        0.001                                   &        -                                 & 54.25                     & 86.18                     & 82.60                     & 81.70                     & \textbf{71.20}            & \textbf{81.23}            & 23.45                     & 95.87                     & 1.05                      & 99.69                     & 46.51                     & 88.93                     & 76.40                         \\
& 256                        &        -                                   &        1                                 & 60.05                     & 84.52                     & 61.75                     & 88.30                     & 78.15                     & 76.60                     & 34.80                     & 93.18                     & 1.50                      & 99.69                     & 47.25                     & 88.46                     & 77.49 \\
\rowcolor[HTML]{EFEFEF}
& 128 &            -                               &      -                                   & \textbf{52.55}            & 87.44                     & 73.95                     & 80.05                     & 71.9                      & 81.17                     & 17.7                      & 97.07                     & 1.3                       & 99.73                     & 43.48                     & 89.09                     & 76.72 \\
\rowcolor[HTML]{EFEFEF}
& 128 & 0.1                                       &        -                                 & 56.45                     & \textbf{87.73}            & 67.45                     & 87.53                     & 78.45                     & 77.24                     & 29.6                      & 94.81                     & 1.95                      & 99.5                      & 46.78                     & 89.36                     & 76.21 \\
\rowcolor[HTML]{EFEFEF}
& 128 &      -                                     & 0.01                                    & 56.05                     & 86.35                     & 83.1                      & 80.75                     & 75.95                     & 79.81                     & 23.5                      & 95.66                     & 1.45                      & 99.65                     & 48.01                     & 88.44                     & 76.35 \\
& 100                        &   -                                        &      -                                   & 57.55                     & 86.8                      & \textbf{54.45}            & \textbf{88.69}            & 75.8                      & 78.88                     & 24.45                     & 95.86                     & 1.6                       & 99.65                     & \textbf{42.77}            & \textbf{89.98}            & 76.41                         \\
& 100                        & 0.001                                     &       -                                  & 82.6                      & 62.38                     & 90.9                      & 60.9                      & 84.4                      & 69.39                     & 67.05                     & 72.8                      & 5.8                       & 97.81                     & 66.15                     & 72.66                     & 31.76                         \\
& 100                        &    -                                       & 1                                       & 100.00                    & 50.00                     & 100.00                    & 50.00                     & 100.00                    & 50.00                     & 100.00                    & 50.00                     & 100.00                    & 50.00                     & 100.00                    & 50.00                     & 1.00 \\ \hline
\end{tabular}%
}
\renewcommand{\arraystretch}{1}
\vspace{-0.1cm}
\end{table*}
\begin{table*}[t]
\centering
\caption{ImageNet-100 Results (ID Acc. $=$ in-distribution accuracy; Null Space Reduction (NSR) methods $=$ our approach).}
\vspace{-3mm}
\label{table:in-100-eval}
\resizebox{0.95\linewidth}{!}{%
\renewcommand{\arraystretch}{1}
\begin{tabular}{lcccccccccccccccc}
\hline
& \multicolumn{3}{c}{\multirow{2}{*}{FEVER-OOD}} &\multicolumn{10}{c}{OOD Datasets}&\\
\cline{5-14}

 &&&& \multicolumn{2}{c}{iNaturalist}& \multicolumn{2}{c}{Places365}& \multicolumn{2}{c}{SUN}& \multicolumn{2}{c}{Textures}&  \multicolumn{2}{c}{Avg}&  \\

 \cline{5-14}

\multirow{-3}{*}{Method}         & $r$-NSR & $\lambda_\mathit{LSV}$ & $\lambda_\mathit{CN}$ & FPR95 & AUROC & FPR95 & AUROC & FPR95 & AUROC & FPR95 & AUROC & FPR95 & AUROC  &        \multirow{-3}{*}{ID Acc}                  \\

\hline\rowcolor[HTML]{EFEFEF}
Dream-OOD&           -                   &              -                             &           -                              & 21.5                         & 96.35                        & 44.60                        & 91.85                        & 42.00                        & 92.45                        & 51.80                        & 85.89                        & 39.98                        & 91.64                        & 87.96                                        \\
\rowcolor[HTML]{EFEFEF}
&            -                   & 0.01                                      &         -                                & 60.10                        & 91.14                        & 66.50                        & 87.95                        & 71.90                        & 85.16                        & 65.10                        & 83.54                        & 65.90                        & 86.95                        & 87.92                                        \\
\rowcolor[HTML]{EFEFEF}
&                      -         &             -                              & 0.001                                   & 60.50                        & 90.76                        & 64.30                        & 88.66                        & 73.30                        & 84.80                        & 65.05                        & 86.90                        & 65.79                        & 87.78                        & 87.68                                        \\
&                     256                           &    -                                       &       -                                  & 22.10                        & 95.95                        & 46.00                        & 91.42                        & 43.50                        & 92.35                        & 45.30                        & 88.11                        & 39.23                        & 91.96                        & 87.30                                        \\
&                     256                           & 0.01                                      &       -                                  & 26.30                        & 95.67                        & 50.20                        & 91.62                        & 43.30                        & 92.21                        & 41.40                        & 89.18                        & 40.30                        & 92.17                        & 87.10                                        \\
&                     256                           &    -                                       & 0.001                                   & 23.40                        & 96.06                        & 48.60                        & 91.23                        & 46.40                        & 91.91                        & 41.10                        & 89.64                        & 39.88                        & 92.21                        & 87.42                                        \\
\rowcolor[HTML]{EFEFEF}
&128 & {\color[HTML]{000000} }         -          & {\color[HTML]{000000} }   -              & {\color[HTML]{000000} 25.40} & {\color[HTML]{000000} 95.99} & {\color[HTML]{000000} 43.70} & {\color[HTML]{000000} 92.19} & {\color[HTML]{000000} 42.30} & {\color[HTML]{000000} 92.70} & {\color[HTML]{000000} 47.00} & {\color[HTML]{000000} 88.81} & {\color[HTML]{000000} 39.60} & {\color[HTML]{000000} 92.42} & {\color[HTML]{000000} 87.50}                 \\
\rowcolor[HTML]{EFEFEF}
&128 & {\color[HTML]{000000} 0.01}               & {\color[HTML]{000000} }    -             & {\color[HTML]{000000} 24.50} & {\color[HTML]{000000} 95.72} & {\color[HTML]{000000} 43.80} & {\color[HTML]{000000} 91.87} & {\color[HTML]{000000} 41.10} & {\color[HTML]{000000} 92.83} & {\color[HTML]{000000} 43.40} & {\color[HTML]{000000} 89.02} & {\color[HTML]{000000} 38.20} & {\color[HTML]{000000} 92.36} & {\color[HTML]{000000} 87.64}                 \\
\rowcolor[HTML]{EFEFEF}
&128 & {\color[HTML]{000000} }          -         & {\color[HTML]{000000} 0.001}            & {\color[HTML]{000000} 23.50} & {\color[HTML]{000000} 96.20} & {\color[HTML]{000000} 42.50} & {\color[HTML]{000000} 91.97} & {\color[HTML]{000000} 43.40} & {\color[HTML]{000000} 92.30} & {\color[HTML]{000000} 42.40} & {\color[HTML]{000000} 89.57} & {\color[HTML]{000000} 37.95} & {\color[HTML]{000000} 92.51} & {\color[HTML]{000000} 87.52}                 \\
& 100                                               &         -                                  &   -                                      & 23.70                        & 95.77                        & 41.60                        & 92.36                        & \textbf{36.60}               & \textbf{93.01}               & \textbf{40.50}               & 89.43                        & \textbf{35.60}               & 92.64                        & 87.36                                        \\
& 100                                               & 0.01                                      &  -                                       & \textbf{20.80}               & \textbf{96.40}               & \textbf{40.00}               & \textbf{92.83}               & 40.60                        & 92.93                        & 42.30                        & \textbf{90.31}               & 35.93                        & \textbf{93.12}               & 87.44                                        \\
&  100                                              &          -                                 & 0.001                                   & 21.60                        & 96.03                        & 42.10                        & 92.69                        & 40.90                        & 92.65                        & 43.70                        & 89.27                        & 37.08                        & 92.66                        & 87.62                                        \\ \hline
\end{tabular}%
}
\renewcommand{\arraystretch}{1}
\vspace{-5mm}
\end{table*}
In this section we discuss the performance of FEVER-OOD against different free energy based OOD detection models in classification and detection.

\noindent
\textbf{CIFAR-10}. \cref{table:cifar-10-eval} shows the results of the VOS and FFS architectures with CIFAR-10 as in-distribution. The reported values for $\lambda_\mathit{LSV}$ (\cref{eq:lsvv-reg}) and $\lambda_\mathit{CN}$ (\cref{eq:cn-reg}) are for the best obtained models (extensive ablations are available in the Supp. Mat.). In general, it is seen that NSR of the last linear layer improves OOD detection for both VOS and FFS models. When evaluating the performance of the NSR with no LSVR or CNR, the average AUROC increases to 93.98 and the average FPR95 reduces to 29.89\% for VOS-96-NSR, compared to an AUROC of 92.17 and 33.3\% of the baseline VOS model without affecting the in-distribution accuracy. Similarly, the overall AUROC and FPR95 for FFS-96-NSR are 94.22 and 27.96\% compared to the baseline FFS model, with an AUROC of 91.57 and FPR95 of 33.96\%. While improvement is also noted for most of the other NSR (without further regularisation), it is also seen that \textit{large} NSR of the last layer might impact OOD detection performance, with increases of the average FPR95 for \{VOS,FFS\}-\{32,10\}-NSR (and AUROC reduction for FFS-10-NSR). Since VOS and FFS are techniques that synthesize outliers in the feature space, a large reduction of the feature dimension might not allow for proper in-distribution understanding, impacting in the OOD training of these techniques. With regards to the LSVR and CNR, it is seen that the LSVR increases the AUROC and reduces the FPR95 in most cases, while CNR can be somewhat unstable with mixed results in both VOS and FFS models. While LSVR increases the difference in energy of outliers in or near the least singular vector direction, the CN regulariser makes equalized free energy changes for all directions. Since the latter effect might disable directions with smaller energy changes than others, it might also reduce the free energy change in the direction of the largest singular value. Overall, the best model for the VOS and FFS architectures are VOS-96-NSR with LSVR, with an avg. FPR95 of 25.48\% (-7.82\% VOS) and AUROC of 95.36 (+3.19 VOS), and FFS-96-NSR with LSVR, with average FPR95 of 24.93\% (-9.03\% FFS) and AUROC of 95.26 (+3.69 FFS).

\begin{table}[t]
\centering
\caption{VOS results with PASCAL VOC as in-distribution \\ (Null Space Reduction (NSR) methods $=$ our approach).}
\vspace{-0.1cm}
\label{table:vos-det-eval}
\resizebox{\linewidth}{!}{%
\renewcommand{\arraystretch}{1}
\begin{tabular}{lcccccccc}
\hline
                                   &                                    \multicolumn{3}{c}{\multirow{2}{*}{FEVER-OOD}}                                         & \multicolumn{4}{c}{OOD Datasets}                                                                                      & \multicolumn{1}{l}{}                      \\ \cline{5-8}
&                                   &                                           &                                         & \multicolumn{2}{l}{MS-COCO}                               & \multicolumn{2}{l}{OpenImages}                            & \multicolumn{1}{l}{}                      \\ \cline{5-8}
\multicolumn{1}{c}{\multirow{-3}{*}{Method}}           & $r$-NSR & $\lambda_\mathit{LSV}$ & $\lambda_\mathit{CN}$ & \multicolumn{1}{l}{FPR95} & \multicolumn{1}{l}{AUROC}     & \multicolumn{1}{l}{FPR95}     & \multicolumn{1}{l}{AUROC} & \multicolumn{1}{l}{\multirow{-3}{*}{mAP}} \\ \hline
\rowcolor[HTML]{EFEFEF}
VOS & -        &               -                            &  -                                       & 50.29                     & 87.77                         & 53.09                         & \textbf{86.58}            & 0.489                                     \\
\rowcolor[HTML]{EFEFEF}
&  -      &    0.001                                   &    -                                     & 49.96                    & 88.10                         & 52.26                         & 85.80                     & 0.489                                     \\
\rowcolor[HTML]{EFEFEF}
&   -     &      -                                     &     1                                    &       53.42                    &          86.66                     &     52.73                          &       85.77                    &     0.490                                      \\
&768                        &   -                                        & -                                        & 48.94                     & 87.94                         & 52.34                         & 86.33                     & 0.489                                     \\
&768                         & 0.01                                      & -                                        & 50.08                     & 87.78                         & 51.71                         & 86.04                     & 0.493                                     \\
&768                         &   -                                        & 0.01                                    & \textbf{47.47}            & 88.15                         & \textbf{49.36}                & 86.32                     & 0.491                                     \\
\rowcolor[HTML]{EFEFEF}
&512 &    -                                       &      -                                   & 49.84                     & \cellcolor[HTML]{EFEFEF}88.42 & \cellcolor[HTML]{EFEFEF}55.33 & 85.33                     & 0.486                                     \\
\rowcolor[HTML]{EFEFEF}
&512 & 0.001                                     &      -                                   & 49.26                     & \cellcolor[HTML]{EFEFEF}88.39 & \cellcolor[HTML]{EFEFEF}54.32 & 85.82                     & 0.489                                     \\
\rowcolor[HTML]{EFEFEF}
&512 &    -                                       & 0.01                                    & 49.59                     & \cellcolor[HTML]{EFEFEF}87.48 & \cellcolor[HTML]{EFEFEF}54.81 & 85.48                     & 0.490                                     \\
&256                         &         -                                 &    -                                     & 47.88                     & \textbf{88.49}                & 52.41                         & 86.39                     & 0.486                                     \\
&256                         & 0.01                                      &     -                                    & 48.51                     & 88.36                         & 50.52                         & 86.12                     & 0.491                                     \\
&256                         &        -                                   & 0.001                                   & 49.1                      & 88.27                         & 51.47                         & 86.08                     & 0.489                                     \\ \hline

\rowcolor[HTML]{EFEFEF}
FFS &      -  &    -                                       &  -                                       & 50.77                     & 87.18                     & 53.78                    & 85.34                  &  0.487                                         \\
\rowcolor[HTML]{EFEFEF}
& -       & 0.01                                           &              -                           &     53.09                      &     86.75                      &     53.47                      &    85.86                       &      0.490                                     \\
\rowcolor[HTML]{EFEFEF}
&  -      &    -                                       &  0.01                                       &    50.36                  & 87.65                     &   51.19                   &   85.98                   &  0.485                                         \\
&768                         &    -                                       &      -                                   &       51.50                    &       87.59                    &      53.33                     &    85.84                       &     0.488                                      \\
&768                         &  0.01                                         &   -                                      &       47.42                    &       88.36                    & 52.46                          &  86.40                         &  0.489                                         \\
&768                         &     -                                      &  0.01                                       & 51.12                 &   86.82                        &  55.44               &    84.84                       &           0.490                                \\
\rowcolor[HTML]{EFEFEF}
&512 & - &- & \textbf{46.93}                                     & \textbf{88.82}                                        &  53.89                         & 86.28                           &  0.491                                         \\
\rowcolor[HTML]{EFEFEF}
&512 &     0.01                                      &     -                                    &     47.42                      & 88.36  & 52.46  &    86.40                       &  0.488                                        \\
\rowcolor[HTML]{EFEFEF}
&512 &       -                                    &       0.01                                  &   51.42                        & 87.05  & 54.98  & 84.55                          &   0.490                                        \\
&256                         &     -                                      &    -                                     &  51.18                         & 88.74                 &  53.67                         &  \textbf{86.56}                         &  0.487                                         \\
&256                         &      0.01                                     &   -                                      &  47.93                         &   88.04                        &   \textbf{47.93}                        &  84.95                         &  0.489                                         \\
&256                         &      -                                     &  0.01                                       &  52.94                         &  86.62                         &  52.88                         &    84.45                       &   0.489                                        \\ \hline

\end{tabular}%
}
\renewcommand{\arraystretch}{1}
\vspace{-3mm}
\end{table}

\noindent
\textbf{CIFAR-100}. \cref{table:cifar-100-eval} shows the results for VOS, FFS and Dream-OOD taking the CIFAR-100 dataset as in-distribution. Whilst the same analysis holds regarding LSVR \textit{vs.} CNR in VOS and FFS as in CIFAR-10, it is observed that the NSR alone has less impact in OOD detection. Since we are using a model with a 128-dimensional feature space (\cref{sec:exp-implementation}), the nullity of the original last layer 18 for CIFAR-100 \textit{vs.} 118 for CIFAR-10 (considering $\mathrm{rank}(\mathrm{W}_\mathit{cls})=K$, where $K$ is the number of classes). Therefore, there is a lower chance to be affected by NSV. The best models for VOS and FFS are achieved with 114-NSR and LSVR, with an 63.51\% FPR95 (-8.93\% VOS) and 84.70 (+3.92 VOS) AUROC for VOS-114-NSR and 65.60\% FPR95 (-6.99\% FFS) and 84.25 (+4.45 FFS) AUROC for FFS-114-NSR. While LSVR and CNR improve OOD performance in Dream-OOD, it is noted that the best results are generally obtained with NSR alone, with Dream-OOD-100-NSR having an FPR95 of 42.77\% (-8.19\% Dream-OOD) and 89.98 AUROC (+1.9 Dream-OOD).
\begin{figure*}[ht]
    \centering
    \includegraphics[width=0.97\linewidth]{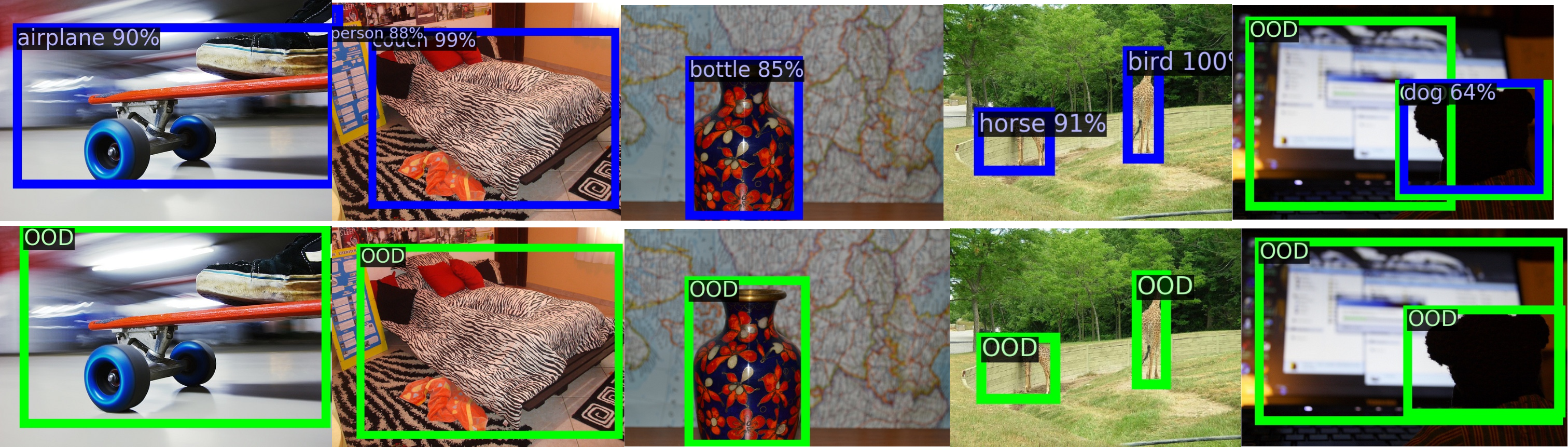} 
    \vspace{-2mm}
    \caption{MS-COCO objects detected on OOD images by VOS baseline \cite{duvos} (first row) and Fever-OOD (second row). \textbf{\textcolor{blue}{Blue}}: OOD objects detected and mis-classified as being in-distribution. \textbf{\textcolor{green}{Green}}: the same OOD objects correctly detected as OOD by FEVER-OOD (ours).}
    \label{fig:placeholder4}
    \vspace{-3mm}
\end{figure*}

\vspace{-0.1cm}
\noindent
\textbf{Imagenet-100}. We show the results of Dream-OOD with Imagenet-100 as in-distribution in \cref{table:in-100-eval}. Similar as with CIFAR-100, the best results are obtained with NSR100, effectively eliminating the null space of the last layer (since it becomes square and is not singular). Contrary to CIFAR-100, LSRV have a negative impact when used alone. Considering that we use the same training scheme as Du \etal \cite{du2023dream} of pre-training Dream-OOD, adding LSVR and CNR on an already pretrained last linear layer might affect the already formed feature space. Hence, the benefit of these regularisers is seen when applying them in untrained NSR modules. The best model, Dream-OOD-100-NSR with LSVR, achieves an average FPR95 of 35.93\% and 93.12 AUROC, compared to the baseline model having 39.98\% FPR95 and 91.64 AUROC, becoming the state-of-the-art for Imagenet-100 OOD detection.


\noindent
\textbf{PASCAL VOC.} We test FEVER-OOD techniques in object detection task, where the goal is to detect object-level OOD instances. \cref{table:vos-det-eval} show the result for the baseline VOS and FFS, taking PASCAL VOC \cite{voc_dataset} as in-distribution dataset. For both tasks, we evaluate on two
OOD datasets that contain subset of images from: MS-COCO \cite{coco_dataset} and OpenImages \cite{kuznetsova2020open}, following the original works \cite{duvos,kumar2023normalizing}. The best-performing model for VOS is VOS-768-NSR with CNR, achieving average FPR95 scores of 47.47\% and 49.36\% on MS-COCO and OpenImages, respectively, an improvement of nearly 4\% over the VOS baseline. For FFS, we achieve a minimum FPR95 of 46.93\% with FFS-512-NSR model and 47.93\% for the FFS-256-NSR model with the LSVR for both OOD datasets. Additionally, our approach generally delivers higher AUROC scores compared to the FFS baseline across both OOD datasets.
Qualitative results are illustrated in \cref{fig:placeholder4}, where we
visualize the prediction on MS-COCO OOD images, using the baseline VOS (top) and the best Fever-OOD model, \ie, VOS-768-NSR, with the PASCAL-VOC as the in-distribution dataset, showing that out framework outperforms baseline VOS in in detecting outlier objects with a significantly lower false positive rate.

\vspace{-0.2cm}
\section{Conclusion}
\label{sec:conclusions}
In this work, we present \textit{FEVER-OOD}, a robust approach for addressing critical vulnerabilities found in existing energy-based OOD detection. Our method targets the Null Space Vulnerabilities (NSV), where in-distribution and OOD instances can produce similar free energy scores despite differing feature representations, and Least Singular Value Vulnerabilities (LSVV), where energy score similarities are influenced by the least singular vector of the classifier last linear layer. These vulnerabilities are especially problematic in high-dimensional feature spaces where the feature dimension exceeds the number of classes, impacting the effectiveness of OOD detection. 
To mitigate these issues, FEVER-OOD incorporates strategies such as reducing the null space through lower-dimensional feature spaces and introducing a Least Singular Value Regularizer to maximize the least singular value, enhancing the distinction between energy scores of different samples. Additionally, we propose a Condition Number Regularizer to promote uniform energy changes for feature-space displacements, fostering consistent energy variation and improving detection robustness.
Our comprehensive evaluation on benchmark datasets, including applications to the Dream-OOD framework with ImageNet-100 as the ID dataset, demonstrates the significant effectiveness of FEVER-OOD. Specifically, our approach achieves a 10.13\% reduction in the average false positive rate and a 1.6\% increase in AUROC, setting a new state-of-the-art in OOD detection.
Furthermore, our results highlight the generalization capability of FEVER-OOD across different detection architectures, showcasing its potential beyond conventional classification settings. Future extensions of FEVER-OOD could include applications to varied domains and integration with other OOD detection frameworks to further validate its versatility and performance in diverse machine learning contexts. This work provides valuable insights into enhancing OOD robustness and sets a foundation for more resilient free energy-based detection methods.
{
    \vspace{-\baselineskip} 
    \small
    \bibliographystyle{ieeenat_fullname}
    \bibliography{main}
}

\clearpage
\maketitlesupplementary

%

\section{Mathematical Background}
In this section we provide a detailed analysis of the mathematical formulation of our methods, as described in \cref{sec:vulnerability}. Specifically, we analyze the constraints in \cref{eq:min-change-energy,eq:min-last-lin-layer} and the solution to the minimization problem in \cref{eq:smin}.

\subsection{Null Space Conditions}

The singular values of a matrix $A \in \mathbb{R}^{m\times n}$ are defined as the square roots of the eigenvalues of the symmetric matrix $A^\top A\in \mathbb{R}^{n\times n}$. Alternatively, the singular vectors of $A$ can be defined as follows:
\begin{align}
    \mathbf{v}_1 &= \argmax_{\|\mathbf{v}\|_2=1} \|A\mathbf{v}\|_2\,,\\
    \label{eq:vi}
    \mathbf{v}_i &= \argmax_{\substack{\|\mathbf{v}\|_2=1\\ \mathbf{v}\perp \mathrm{span}\{\mathbf{v}_1,\dots,\mathbf{v}_{i-1}\}}} \|A\mathbf{v}\|_2,\quad i\geq 2\,.
\end{align}
The singular values of $A$ are then given by:
\begin{equation}
    \sigma_i(A) = \|A\mathbf{v}_i\|_2\,.
\end{equation}
The null space of $A$, denoted as $\mathrm{Null}(A)$, is defined as the span of unit vectors whose corresponding singular values are zero. From \cref{eq:vi}, there are $n$ singular vectors (as there are $n$ orthogonal vectors in $\mathbb{R}^n$). If $r=\mathrm{rank}(A)$, the rank-nullity theorem implies that the nullity, \ie, the dimension of $\mathrm{Null}(A)$, is $n - r$. Therefore, from \cref{eq:vi}, the null space of $A$ is given by:
\begin{equation} \label{eq:null-space}
    \mathrm{Null}(A) = \mathrm{span}\{\mathbf{v}_{r+1},...,\mathbf{v}_n\}\,.
\end{equation}
Moreover, if $\mathbf{v}_1,\dots,\mathbf{v}_r$ are the singular vectors corresponding to the non-zero singular values of $A$, $\sigma_1(A)\geq \dots\geq \sigma_r(A)>0$ (assuming that $A\neq 0$), then the span of these vectors is orthogonal to the null space of $A$. Formally:
\begin{equation} \label{eq:perp-null-space}
\mathrm{Null}(A)^\perp = \mathrm{span}\{\mathbf{v}_1,\dots,\mathbf{v}_r\}\,.
\end{equation}
Since $\mathbf{v}_1,\hdots,\mathbf{v}_n$ are an orthonormal basis of $\mathbb{R}^n$, and the subspaces in \cref{eq:null-space,eq:perp-null-space} are complementary, every feature vector $\boldsymbol{\nu}\in\mathbb{R}^n$ can be written as:
\begin{equation} \label{eq:space-decomposition}
    \boldsymbol{\nu}=\boldsymbol{\nu}_n+\boldsymbol{\nu}_a\,,\ \boldsymbol{\nu}_n\in\mathrm{Null}(A)\,,\ \boldsymbol{\nu}_a\in\mathrm{Null}(A)^\perp\,.
\end{equation}
This decomposition indicates that every vector $\boldsymbol{\nu} \in \mathbb{R}^n$ consists of a component $\boldsymbol{\nu}_n$ in the null space of $A$ and a component $\boldsymbol{\nu}_a$ orthogonal to $\mathrm{Null}(A)$. This makes the condition $\boldsymbol{\delta} \in \mathrm{Null}(\mathrm{W}_\mathit{cls}^\top)^\perp$ in \cref{eq:min-change-energy,eq:min-last-lin-layer,eq:smin} necessary to avoid a trivial solution to the minimization problem. For instance, if this condition were not enforced, then the solution would be zero, corresponding to any $\boldsymbol{\delta} \in \mathrm{Null}(\mathrm{W}_\mathit{cls}^\top)$. If we only enforced $\boldsymbol{\delta} \notin \mathrm{Null}(\mathrm{W}_\mathit{cls}^\top)$, then the minimum would not exist but the infimum would be zero:
\begin{equation} \label{eq:inf}
\inf_{\substack{\boldsymbol{\delta}: \lVert \boldsymbol{\delta} \rVert \ge d_b,\\\boldsymbol{\delta} \notin \mathrm{Null}(\mathrm{W}^\top_\mathit{cls})}} \lVert \mathrm{W}^\top_\mathit{cls}\boldsymbol{\delta}\rVert=0\,.
\end{equation}
\cref{eq:inf} holds because $\boldsymbol{\delta}$ can be decomposed into components $\boldsymbol{\delta}_n \in \mathrm{Null}(A)$ and $\boldsymbol{\delta}_a \in \mathrm{Null}(A)^\perp$, where $\boldsymbol{\delta}_n$ can be arbitrarily larger than $\boldsymbol{\delta}_a$, making $ \lVert \mathrm{W}^\top_\mathit{cls}\boldsymbol{\delta}\rVert$ arbitrarily small (but not zero).

\subsection{Least Singular Value Solution}
From \cref{eq:vi,eq:perp-null-space}, it follows that:
\begin{equation}\label{eq:supp-least singular value}
\min_{\substack{\|\mathbf{v}\|_2=1\\\mathbf{v}\in \mathrm{Null}(A)^\perp}} \|A\mathbf{v}\|_2 = \sigma_r(A) =: \sigma_{min}(A).
\end{equation}
This equation highlights that the smallest non-zero singular value of $A$ corresponds to the minimum norm of $A\mathbf{v}$ over all unit vectors orthogonal to $\mathrm{Null}(A)$.

The singular value decomposition (SVD) encodes information about the singular values and singular vectors of a given matrix in a structured way. Namely, any matrix $A\in \mathbb{R}^{m\times n}$ can be factorized as follows:
\begin{equation}
A=U\Sigma V^\top\,,
\end{equation}
where:
\begin{itemize}
\item $\Sigma\in \mathbb{R}^{m\times n}$ is a diagonal rectangular matrix whose entries are the singular values of $A$ in descending order;
\item $U\in \mathbb{R}^{m\times m}$ and $V\in\mathbb{R}^{n\times n}$ are orthogonal matrices, whose columns correspond to the left and right singular vectors of $A$, respectively.
\end{itemize}
In particular, the SVD of $A^\top$ is given by: 
\begin{equation}
A^\top = V\Sigma^\top U^\top\,,
\end{equation}
which implies that $A$ and $A^\top$ have the same non-zero singular values. Combining this with Equation \eqref{eq:supp-least singular value}, we get
\begin{equation}
\min_{\substack{\|\mathbf{v}\|_2=1\\\mathbf{v}\in \mathrm{Null}(A^\top)^\perp}} \|A^\top\mathbf{v}\|_2 = \sigma_{min}(A)\,,
\end{equation}
which justifies Equation \eqref{eq:smin}.

\section{Training Regime}
We follow the original training regime for each baseline method and their corresponding FEVER-OOD variants\footnote{Code is available at \url{https://github.com/KostadinovShalon/fever-ood}.}.
\paragraph{VOS} \cite{duvos}: we train all our VOS for classification models for 100 epochs with a batch size of 128 $32\times32$ images (CIFAR-10 and CIFAR-100 \cite{krizhevsky2009learning} datasets). Outlier synthesis started in epoch 40 in all experiments. Following Du \etal \cite{duvos}, we sample $10,000$ instances per category in the feature space and choose the instance with the least log probability as the outlier. We use an initial learning rate of 0.1 with cosine annealing and stochastic gradient descent (SGD) with $5\times10^{-4}$ weight decay and 0.9 momentum for all experiments. A loss weight of 0.1 is used for the uncertainty loss. With regards to VOS for detection models, we follow the same outlier synthesis scheme as in classification. We use a batch size of 16 images with varying minimum width from 480 to 800 pixels, and train for $18,000$ iterations, corresponding to around 17.4 epochs for the PASCAL VOC \cite{voc_dataset} dataset. An initial learning rate of 0.02 is used for all VOS detection models, decaying by a factor of 10 after 12,000 epochs and again after 16,000 epochs. Similarly to classification, loss weight of 0.1 us used for the uncertainty loss. All VOS training was carried out using a single GPU per experiment. VOS classification models were trained on NVIDIA GeForce RTX 2080 Ti GPUs, while VOS detection models were trained on NVIDIA RTX A6000 GPUs.
\paragraph{FFS} \cite{kumar2023normalizing}: FFS models follow a similar training regime as VOS models. The only difference for FFS models is that the outliers are obtained as the least likely out of  $200$ samples from the normalizing flow feature space, following Kumar \etal \cite{kumar2023normalizing}. We use the same 0.1 loss weight for the uncertainty loss as in VOS, and $1\times10^{-4}$ loss weight for the normalizing flow loss, for both classification and detection models. Additionally, we implement FFS for classification since the original implementation is only for object classification.
\paragraph{Dream-OOD} \cite{du2023dream}: we train Dream-OOD in CIFAR-100 for 100 epochs with a batch size of 160 in-distribution images and 160 OOD images (for a combined 320 epochs per batch). We use SGD with an initial learning rate of 0.1, cosine annealing, $5\times10^{-4}$ weight decay and 0.9 momentum. Regarding Imagenet-100, we use a ResNet-34 \cite{he2016deep} that is pretrained solely on image classification only for 100 epochs. We train for OOD detection for 20 further epochs, using a batch size of 20 in-distribution and 20 OOD images, initial learning rate of 0.001, $5\times10^{-4}$ weight decay and 0.9 momentum. Following Du \etal \cite{du2023dream}, we use an energy loss weight of 2.5 for CIFAR-100 experiments and 1.0 for Imagenet-100 experiments. Dream-OOD for CIFAR-10 models were trained with single NVIDIA GeForce RTX 2080 Ti GPUs while the Imagenet-100 models were trained with single NVIDIA TESLA v100 GPUs. We use the ResNet-34 pretrained model for Imagenet-100 and the generated outliers in the pixel space for both CIfAR-100 and Imagenet-100 provided by Du \etal \cite{du2023dream} at \url{https://github.com/deeplearning-wisc/dream-ood}. 
\vspace{-0.2cm}

\section{Null Space Projection}
\begin{figure*}[t]
    \centering
    \begin{subfigure}[c]{0.325\textwidth}
        \centering
        \includegraphics[width=\textwidth]{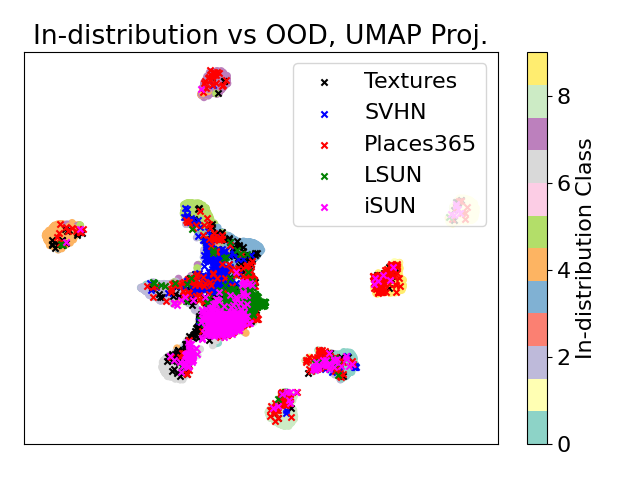}
        \includegraphics[width=\textwidth]{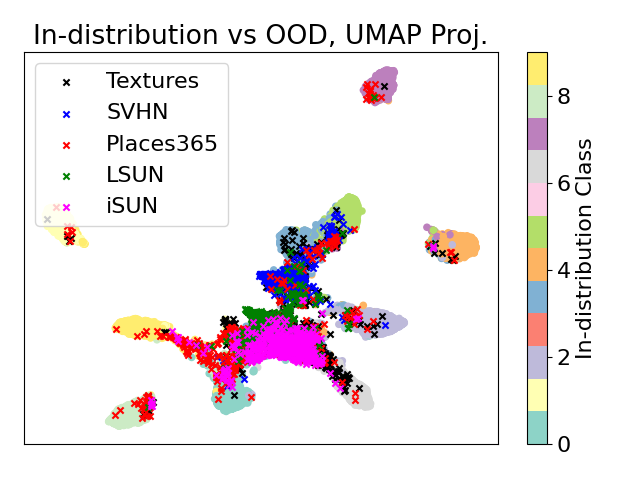}
        \caption{}
        \label{fig:id-ood-vos-proj}
    \end{subfigure}
    \hfill
    \begin{subfigure}[c]{0.325\textwidth}
        \centering
        \includegraphics[width=\textwidth]{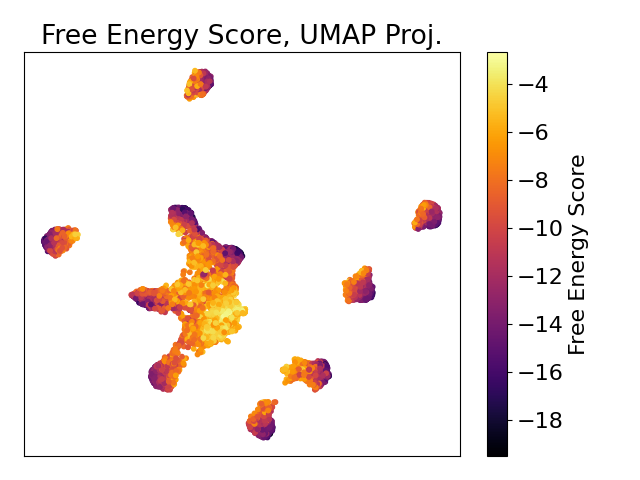}
        \includegraphics[width=\textwidth]{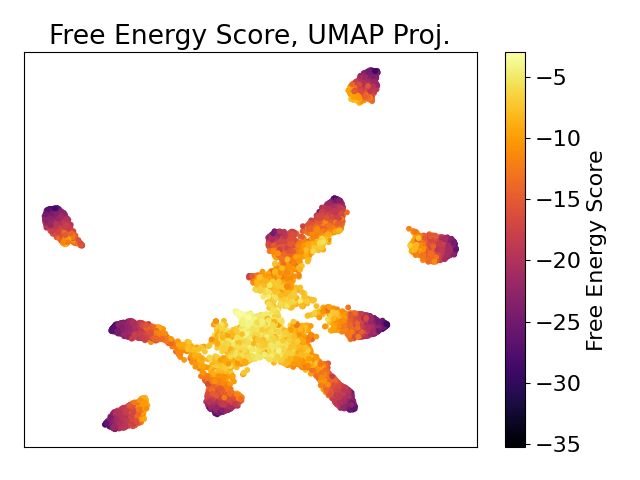}
        \caption{}
        \label{fig:energy-vos-proj}
    \end{subfigure}
    \hfill
    \begin{subfigure}[c]{0.325\textwidth}
        \centering
        \includegraphics[width=\textwidth]{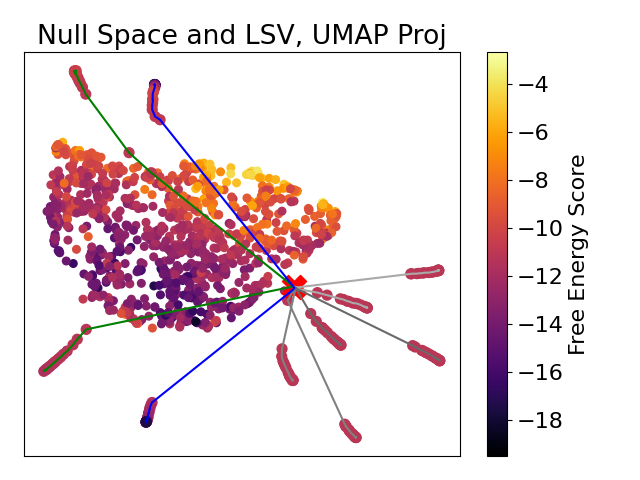}
        \includegraphics[width=\textwidth]{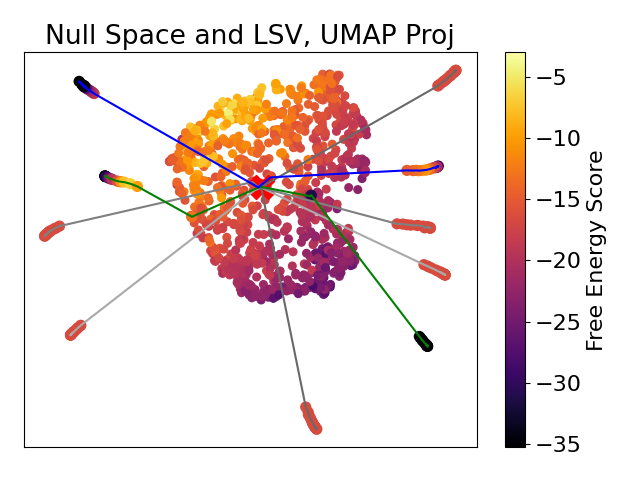}
        \caption{}
        \label{fig:directions-vos-proj}
    \end{subfigure}
    \caption{Feature Space UMAP Projection for models trained on CIFAR-10. Top row corresponds to the VOS \cite{duvos} model while the bottom shows the FEVER-OOD VOS (Ours) projections. (a) In-distribution vs OOD feature space projection, (b) Free Energy visualization of the feature space, and (c) different important directions, including \textcolor{gray}{Null Space directions}, the \textcolor{green}{LSV direction} and a \textcolor{blue}{random direction}.}
    \label{fig:feature-projection-vos}
\end{figure*}

\begin{figure*}[t]
    \centering
    \begin{subfigure}[c]{0.325\textwidth}
        \centering
        \includegraphics[width=\textwidth]{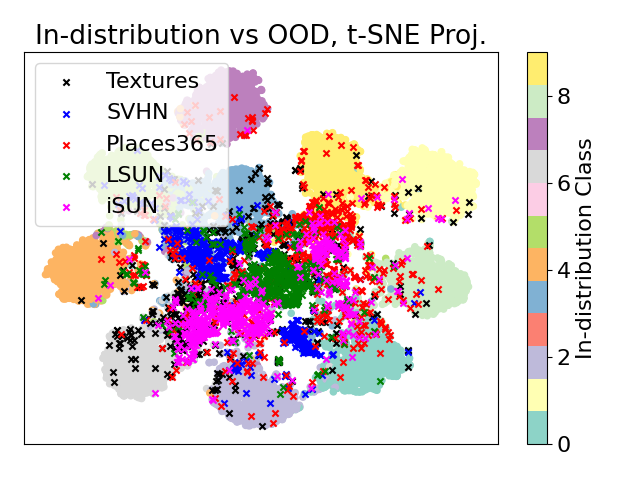}
        \includegraphics[width=\textwidth]{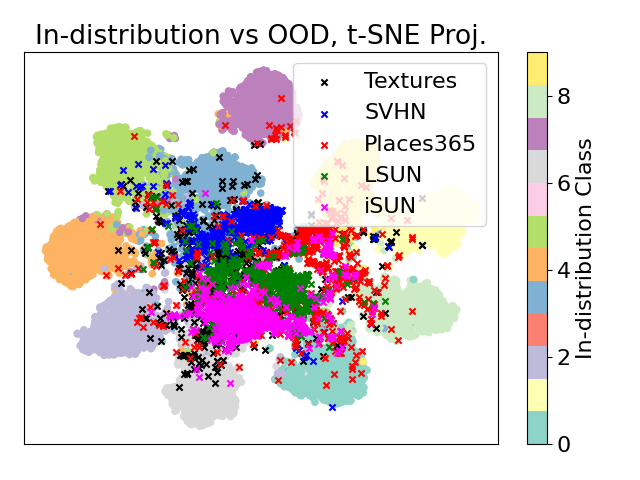}
        \caption{}
        \label{fig:id-ood-fever-vos-proj}
    \end{subfigure}
    \hfill
    \begin{subfigure}[c]{0.325\textwidth}
        \centering
        \includegraphics[width=\textwidth]{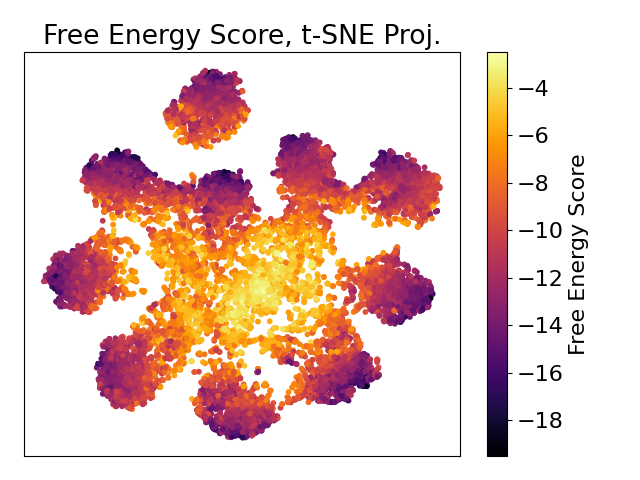}
        \includegraphics[width=\textwidth]{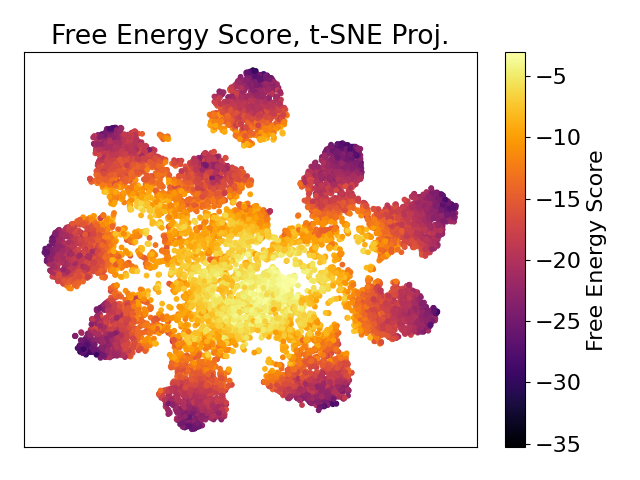}
        \caption{}
        \label{fig:energy-fever-vos-proj}
    \end{subfigure}
    \hfill
    \begin{subfigure}[c]{0.325\textwidth}
        \centering
        \includegraphics[width=\textwidth]{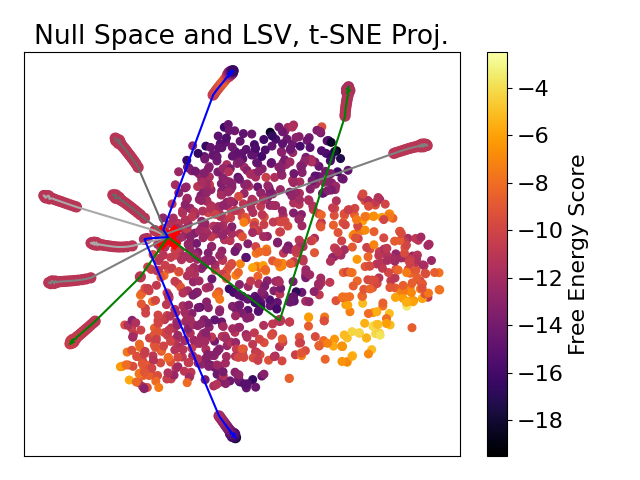}
        \includegraphics[width=\textwidth]{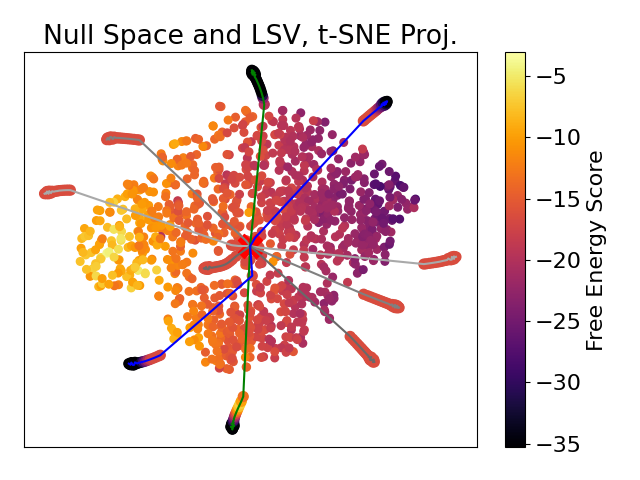}
        \caption{}
        \label{fig:directions-fever-vos-proj}
    \end{subfigure}
    \caption{Feature Space t-SNE Projection for models trained on CIFAR-10. Top row corresponds to the VOS \cite{duvos} model while the bottom shows the FEVER-OOD VOS (Ours) projections. (a) In-distribution vs OOD feature space projection, (b) Free Energy visualization of the feature space, and (c) different important directions, including \textcolor{gray}{Null Space directions}, the \textcolor{green}{LSV direction} and a \textcolor{blue}{random direction}.}
\label{fig:feature-projection-fever-vos}

\end{figure*}

\cref{fig:feature-projection-vos,fig:feature-projection-fever-vos} show the feature projections of VOS and FEVER-OOD VOS models using UMAP and t-SNE projections, respectively. Both models are for CIFAR-10 as in-distribution data, with the best model of FEVER-OOD VOS being shown, corresponding to an 96-NSR and $\lambda_\mathit{LSV}=1.0$ (\cref{table:cifar-10-eval}). Specifically, \cref{fig:id-ood-vos-proj,fig:id-ood-fever-vos-proj} show the projection of in-distribution vs. OOD examples. The projections of the feature space of both methods show that the OOD samples are pushed to different regions outside, with defined clusters for in-distribution classes (colored according to their ground-truth class). Nonetheless, the free energy score for these samples shown in \cref{fig:energy-vos-proj,fig:energy-fever-vos-proj} exhibits a more uniform distribution of the free energy for OOD samples when using FEVER-OOD, providing a visualization of why our technique works better. Additionally, the different scales of the energy values between both models indicates a higher separability between in-distribution and OOD. Finally, \cref{fig:directions-vos-proj,fig:directions-fever-vos-proj} show the feature distribution of the class 0 and some virtually generated features along some directions. Specifically, we generate features away from the centroid of the in-distribution feature vectors along three null space directions (gray), the LSV direction (green) and a random direction (blue). As described in \cref{sec:ns_last_layer}, the energy of the features in the null space direction does not change, which is reflected in \cref{fig:directions-vos-proj,fig:directions-fever-vos-proj}, where all the null space features have the same energy. Additionally, \cref{sec:minimal_energy} shows that the direction with least change in energy corresponds to the LSV direction. In this sense, because our FEVER-OOD VOS shown in \cref{fig:directions-vos-proj,fig:directions-fever-vos-proj} uses the LSV regularizer, the energy plots show greater energy difference in the LSV direction (green line) for our method when compared with VOS. 

\begin{figure*}[t]
    \centering
    \begin{subfigure}[c]{0.42\textwidth}
        \centering
        \includegraphics[width=\textwidth]{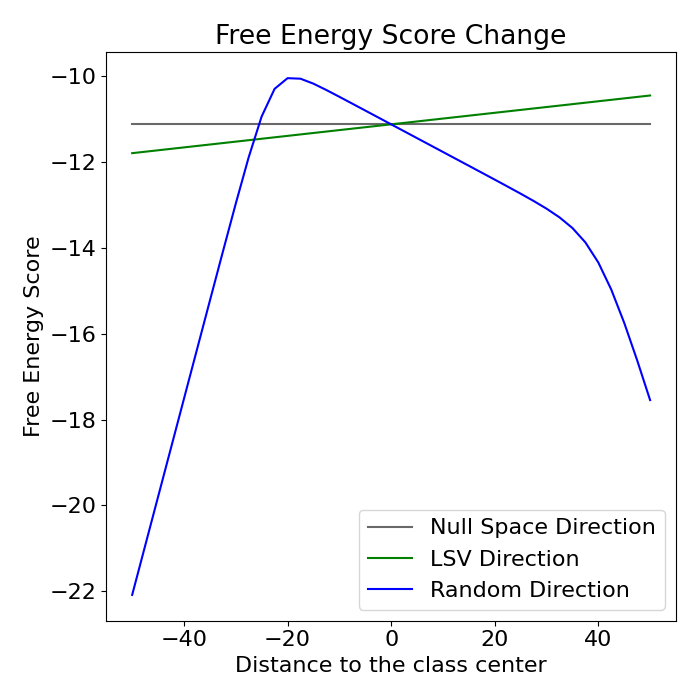}
        \caption{VOS}
    \end{subfigure}
    \hfill
    \begin{subfigure}[c]{0.42\textwidth}
        \centering
        \includegraphics[width=\textwidth]{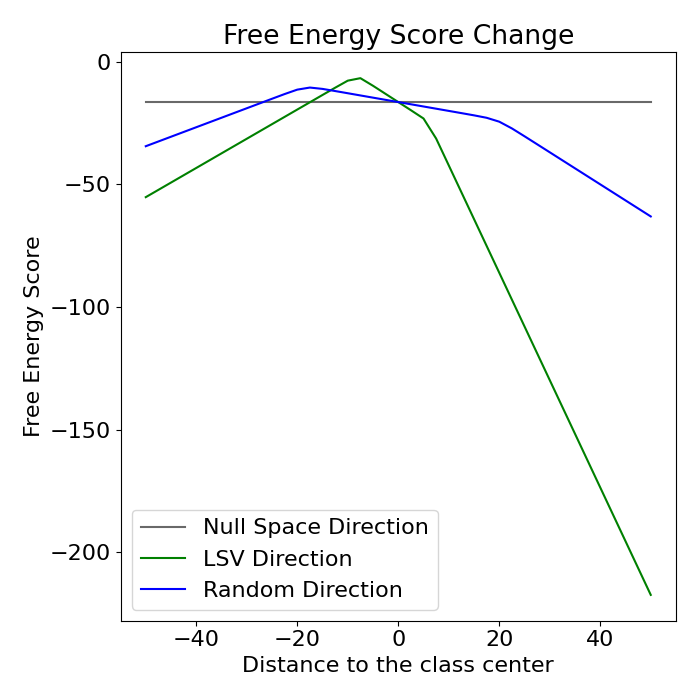}
        \caption{FEVER-VOS}
    \end{subfigure}
    \caption{Free energy change by its distance to the centroid of the feature vectors of an in-distribution category along different directions. (a) VOS vs. (b) FEVER-OOD VOS.}
    \label{fig:free-energy-change-vos-vs-fever-ood-vos}
\end{figure*}
\begin{figure*}[t]
    \centering
    \begin{subfigure}[c]{0.45\textwidth}
        \centering
        \includegraphics[width=\textwidth]{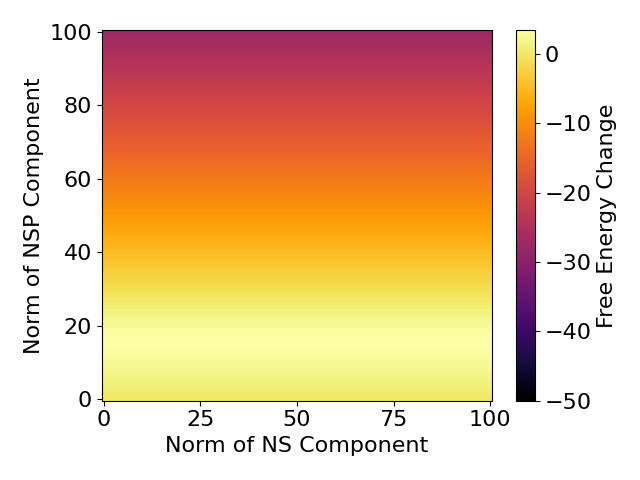}
        \caption{VOS}
    \end{subfigure}
    \hfill
    \begin{subfigure}[c]{0.45\textwidth}
        \centering
        \includegraphics[width=\textwidth]{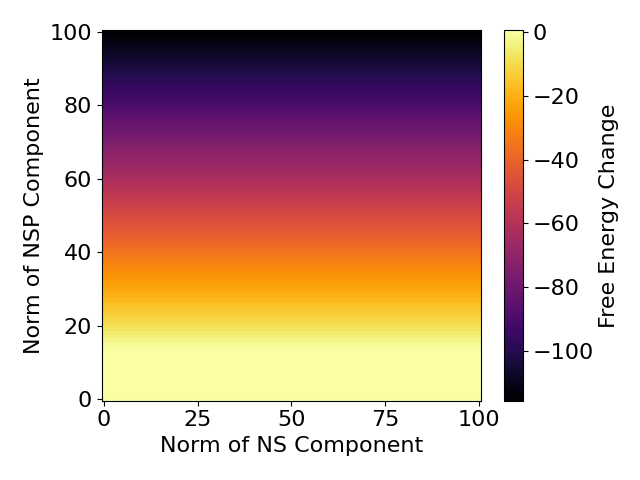}
        \caption{FEVER-VOS}
    \end{subfigure}
    \caption{Free energy change for varying the contribution of the Null Space (NS) component and the Null Space Perpendicular (NSP) component. (a) VOS vs. (b) FEVER-OOD VOS.}
    \label{fig:free-energy-change-components}
\end{figure*}

\cref{fig:free-energy-change-vos-vs-fever-ood-vos} shows the change in energy across these directions with respect to the distance to the in-distribution centre, where it is seen that the change in energy in the LSV direction is significantly larger with FEVER-OOD. Finally, with regards to the component decomposition in \cref{eq:space-decomposition}, \cref{fig:free-energy-change-components} shows the energy change with respect to an in-distribution feature vector when moving in directions with varying contribution of the components of the null space and perpendicular to the null space. Since the null space component does not change the free energy score, all the changes are in the vertical direction, showing a greater change when using FEVER-OOD vs. the baseline methods.

\begin{figure*}[t]
    \centering
    \begin{subfigure}[c]{0.19\textwidth}
        \centering
        \includegraphics[width=\textwidth]{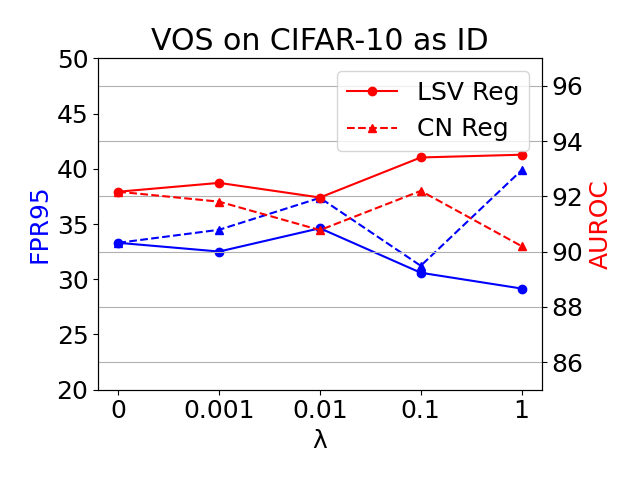}
        \caption{VOS}
    \end{subfigure}
    \hfill
    \begin{subfigure}[c]{0.19\textwidth}
        \centering
        \includegraphics[width=\textwidth]{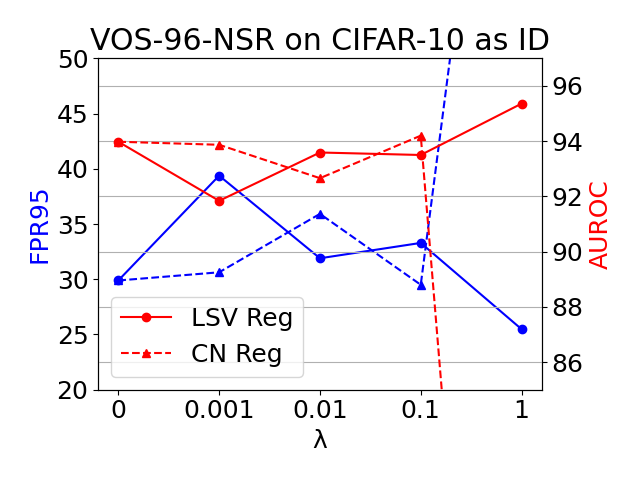}
        \caption{VOS-96-NSR}
    \end{subfigure}
    \hfill
    \begin{subfigure}[c]{0.19\textwidth}
        \centering
        \includegraphics[width=\textwidth]{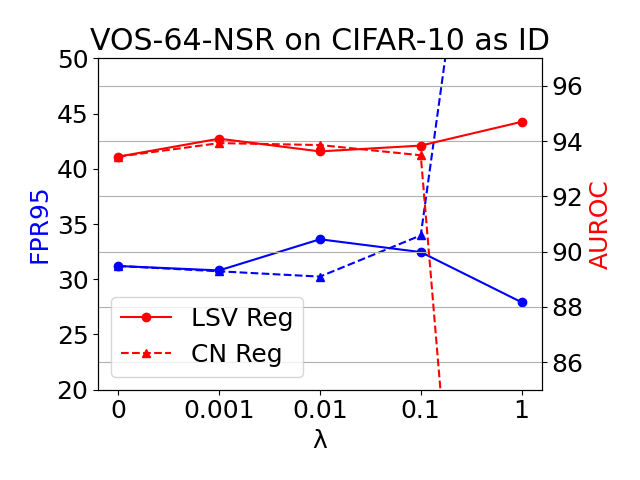}
        \caption{VOS-64-NSR}
    \end{subfigure}
    \hfill
    \begin{subfigure}[c]{0.19\textwidth}
        \centering
        \includegraphics[width=\textwidth]{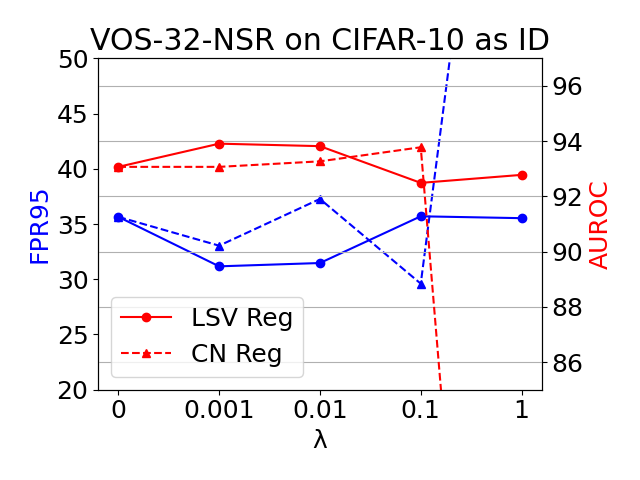}
        \caption{VOS-32-NSR}
    \end{subfigure}
    \hfill
    \begin{subfigure}[c]{0.19\textwidth}
        \centering
        \includegraphics[width=\textwidth]{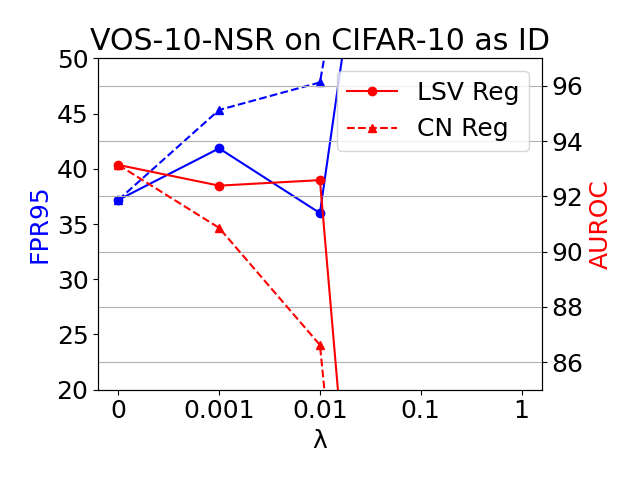}
        \caption{VOS-10-NSR}
    \end{subfigure}
    
    \caption{Ablations of the loss weight for the LSV and CN regularizers in FEVER-OOD for VOS, using CIFAR-10 as in-distribution (ID).}
    \label{fig:vos-cifar-10-ablations}
\end{figure*}

\begin{figure*}[t]
    \centering
    \begin{subfigure}[c]{0.19\textwidth}
        \centering
        \includegraphics[width=\textwidth]{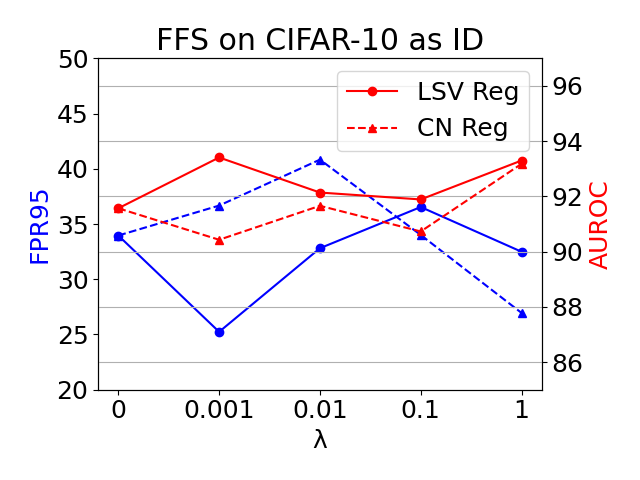}
        \caption{FFS}
    \end{subfigure}
    \hfill
    \begin{subfigure}[c]{0.19\textwidth}
        \centering
        \includegraphics[width=\textwidth]{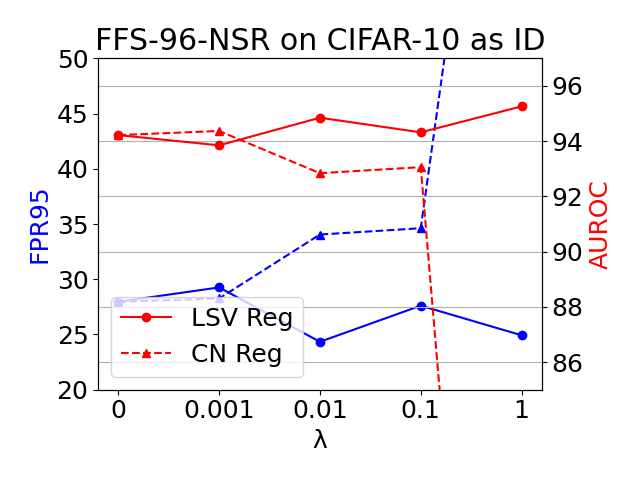}
        \caption{FFS-96-NSR}
    \end{subfigure}
    \hfill
    \begin{subfigure}[c]{0.19\textwidth}
        \centering
        \includegraphics[width=\textwidth]{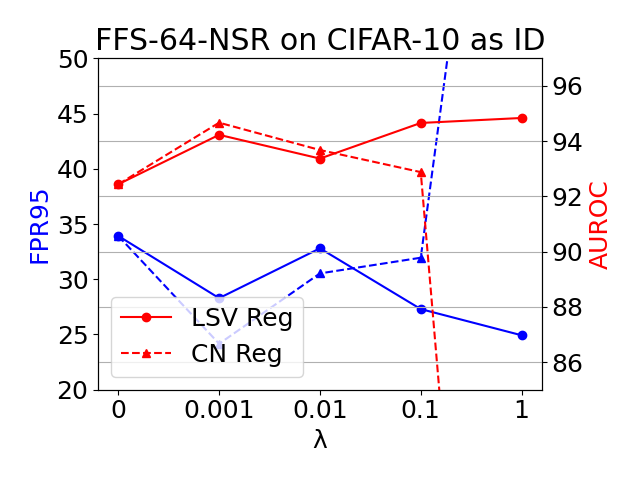}
        \caption{FFS-64-NSR}
    \end{subfigure}
    \hfill
    \begin{subfigure}[c]{0.19\textwidth}
        \centering
        \includegraphics[width=\textwidth]{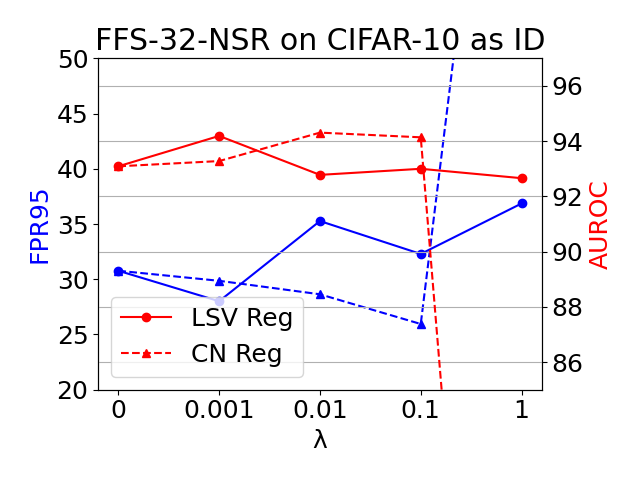}
        \caption{FFS-32-NSR}
    \end{subfigure}
    \hfill
    \begin{subfigure}[c]{0.19\textwidth}
        \centering
        \includegraphics[width=\textwidth]{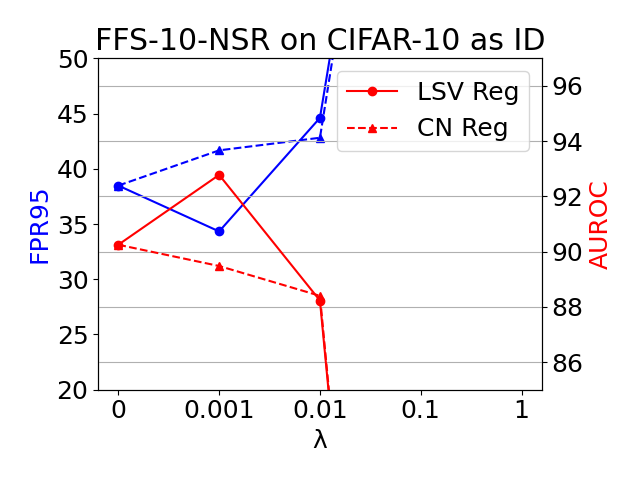}
        \caption{FFS-10-NSR}
    \end{subfigure}
    
    \caption{Ablations of the loss weight for the LSV and CN regularizers in FEVER-OOD for FFS, using CIFAR-10 as in-distribution (ID).}
    \label{fig:ffs-cifar-10-ablations}
\end{figure*}

\begin{figure*}[ht]
    \centering
    \begin{subfigure}[c]{0.30\textwidth}
        \centering
        \includegraphics[width=\textwidth]{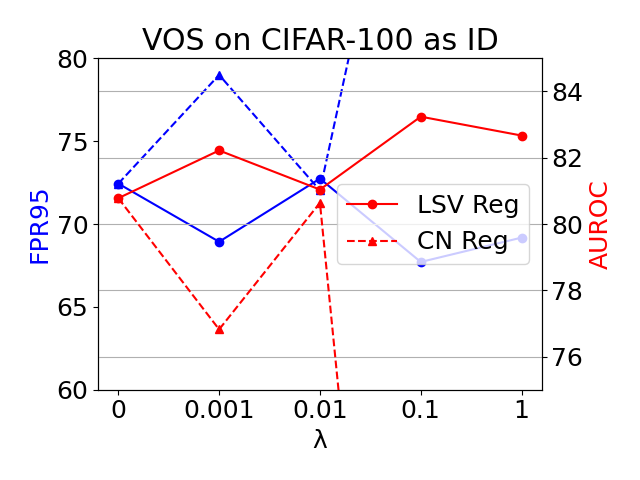}
        \caption{VOS}
    \end{subfigure}
    \hfill
    \begin{subfigure}[c]{0.30\textwidth}
        \centering
        \includegraphics[width=\textwidth]{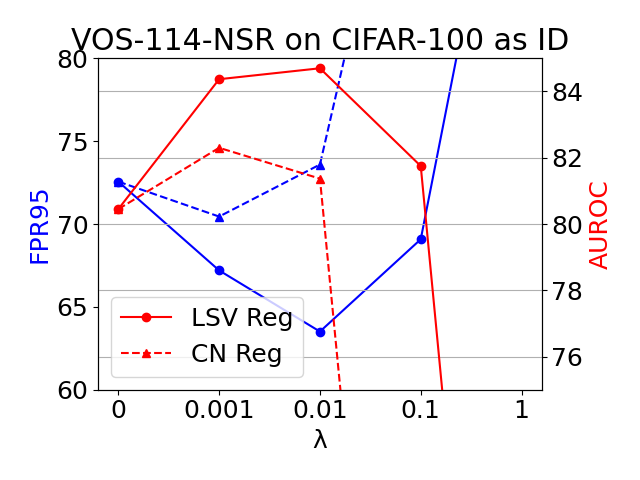}
        \caption{VOS-114-NSR}
    \end{subfigure}
    \hfill
    \begin{subfigure}[c]{0.30\textwidth}
        \centering
        \includegraphics[width=\textwidth]{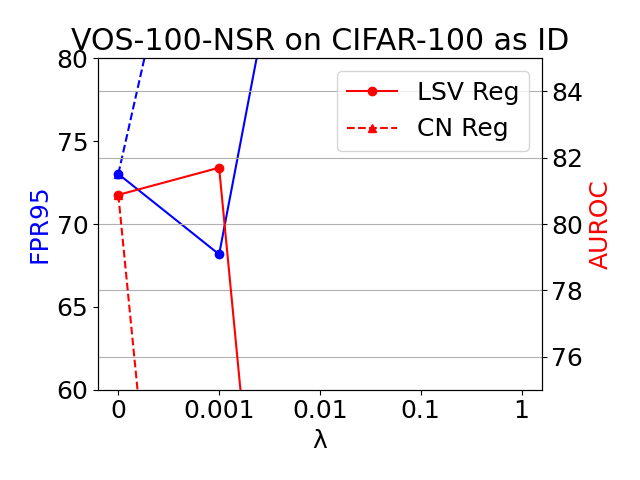}
        \caption{VOS-100-NSR}
    \end{subfigure}
    \caption{Ablations of the loss weight for the LSV and CN regularizers in FEVER-OOD for VOS, using CIFAR-100 as in-distribution (ID).}
    \label{fig:vos-cifar-100-ablations}
\end{figure*}

\begin{figure*}[t]
    \centering
    \begin{subfigure}[c]{0.30\textwidth}
        \centering
        \includegraphics[width=\textwidth]{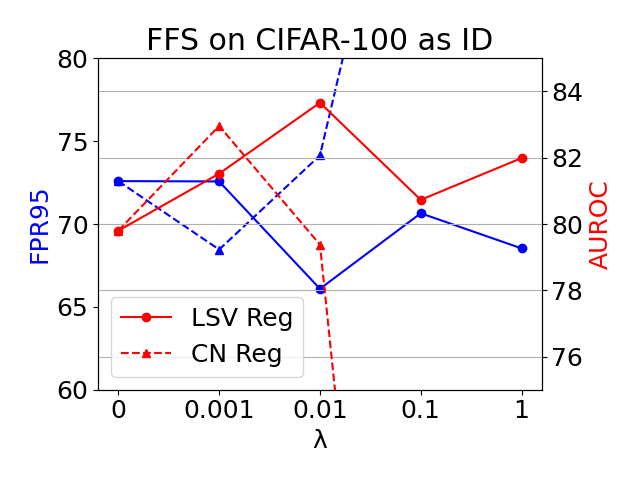}
        \caption{FFS}
    \end{subfigure}
    \hfill
    \begin{subfigure}[c]{0.30\textwidth}
        \centering
        \includegraphics[width=\textwidth]{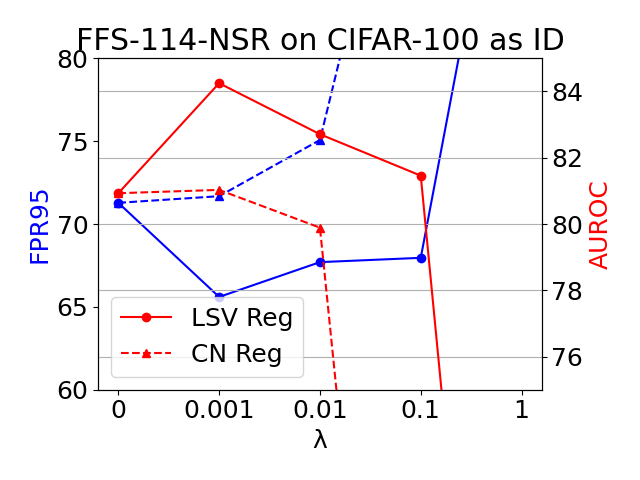}
        \caption{FFS-114-NSR}
    \end{subfigure}
    \hfill
    \begin{subfigure}[c]{0.30\textwidth}
        \centering
        \includegraphics[width=\textwidth]{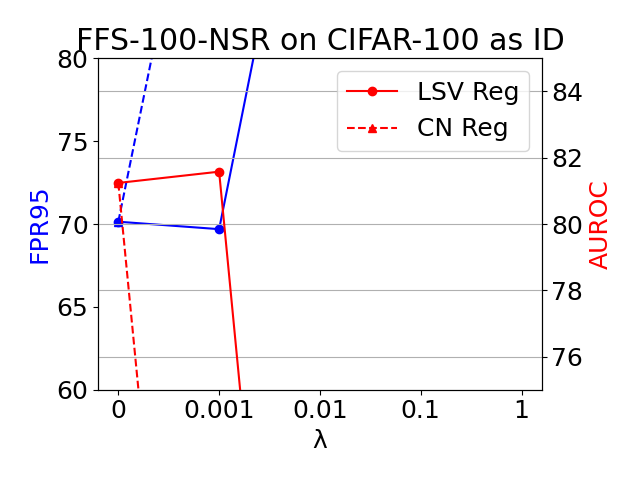}
        \caption{FFS-100-NSR}
    \end{subfigure}
    \caption{Ablations of the loss weight for the LSV and CN regularizers in FEVER-OOD for FFS, using CIFAR-100 as in-distribution (ID).}
    \label{fig:ffs-cifar-100-ablations}
\end{figure*}

\begin{figure*}[t]
    \centering
    \begin{subfigure}[c]{0.22\textwidth}
        \centering
        \includegraphics[width=\textwidth]{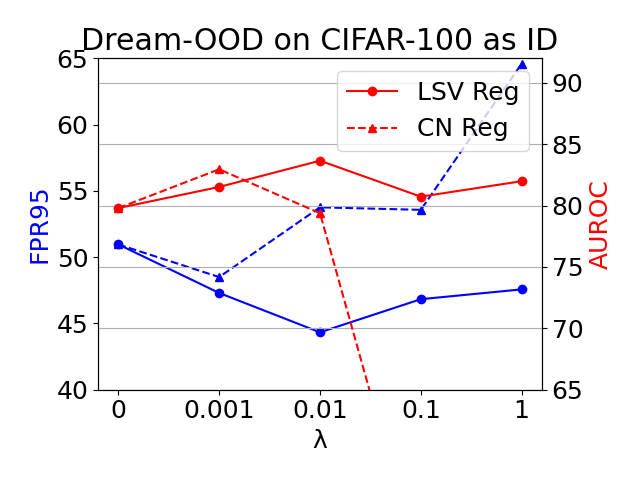}
        \caption{Dream-OOD}
    \end{subfigure}
    \hfill
    \begin{subfigure}[c]{0.22\textwidth}
        \centering
        \includegraphics[width=\textwidth]{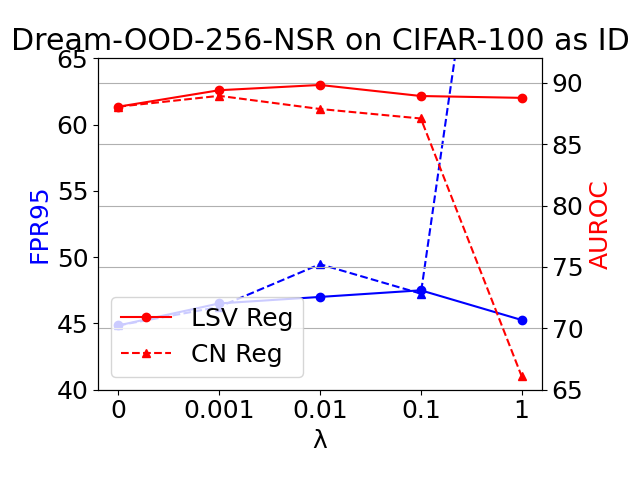}
        \caption{Dream-OOD-256-NSR}
    \end{subfigure}
    \hfill
    \begin{subfigure}[c]{0.22\textwidth}
        \centering
        \includegraphics[width=\textwidth]{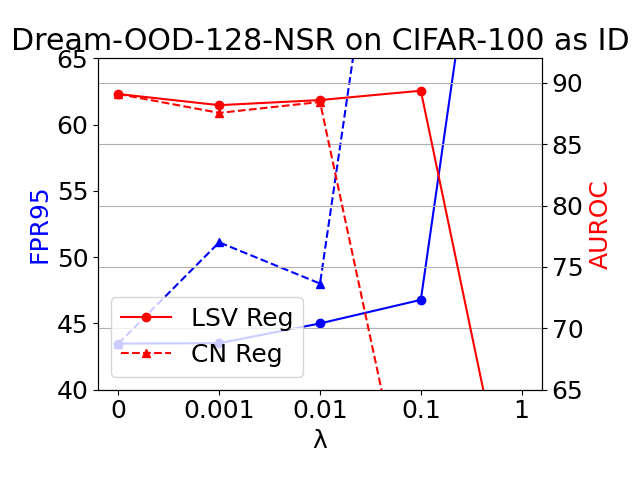}
        \caption{Dream-OOD-128-NSR}
    \end{subfigure}
    \hfill
    \begin{subfigure}[c]{0.22\textwidth}
        \centering
        \includegraphics[width=\textwidth]{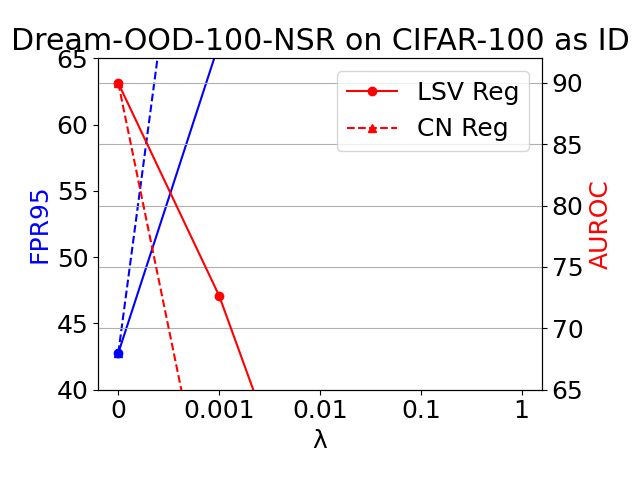}
        \caption{Dream-OOD-100-NSR}
    \end{subfigure}
    \caption{Ablations of the loss weight for the LSV and CN regularizers in FEVER-OOD for Dream-OOD, using CIFAR-100 as in-distribution (ID).}
    \label{fig:dream-ood-cifar-100-ablations}
\end{figure*}

\section{Ablation Studies}
\cref{fig:vos-cifar-10-ablations,fig:vos-cifar-100-ablations,fig:ffs-cifar-10-ablations,fig:ffs-cifar-100-ablations,fig:dream-ood-cifar-100-ablations} show the ablation studies of varying $\lambda_\mathit{LSV}$ and $\lambda_\mathit{CN}$ in \cref{eq:lsvv-reg,eq:cn-reg} for different classification methods and in-distribution datasets. \cref{fig:vos-cifar-10-ablations} shows the results for FEVER-OOD VOS using CIFAR-10 as in-distribution, where values corresponding to a $\lambda_{\{\mathit{LSV}, \mathit{CN}\}}=0$ refer to no LSV or CN regularisation. In general,  LSV regularisation gives better results than CN regularisation. It is observed that larger values of $\lambda_\mathit{LSV}$ leads to better performance for none or few NSR (VOS, VOS-96-NSR and VOS-64-NSR). The same trend is observed for FEVER-OOD FFS in \cref{fig:ffs-cifar-10-ablations}. It is also observed that the models become unstable when using a large NSR, where the extreme case of \{VOS,FFS\}-10-NSR failrs for both regularizer at relative small loss weights. This effect could be caused because regualrizing the least singular value (either for LSVR or CNR) affects all the directions of the feature space since there is no null space. This causes makes the model not able to learn the in-distribution task, failing also for OOD detection. Additionally, all NSR models fail for $\lambda_\mathit{CN}=1.0$. 

FEVER-OOD VOS and FFS ablations for the CIFAR-100 as in-distribution are shown in \cref{fig:vos-cifar-100-ablations,fig:ffs-cifar-100-ablations}. Similar as with CIFAR-10, LSV regularisation is more stable and leads to better results than CN regularisation. In both OOD models (VOS and FFS), it is observed that the best results are achieved with 114-NSR and an intermediate value for $\lambda_\mathit{LSV}$. These results indicate that for OOD models based in outlier generation in the feature space, some reduction of the null space and a moderate regularizer is benefical. However, the complete elimination of the null space and LSV (or CN) regularization might impose a huge prior in the last layer, making it difficult to learn the in-distribution task. Finally, \cref{fig:dream-ood-cifar-100-ablations} shows the ablations for FEVER-OOD Dream-OOD with CIFAR-100 as in-distribution. Here it is observed that NSR by itself leads to better results, suggesting that there might be a significant portion of generated outliers in with large components in the null space of the feature space. Dream-OOD follows a similar patter as the other models for CIFAR-100, suggesting that the analysis holds for different OOD approaches.

\section{Qualitative Results}
\begin{figure*}[t!]
    \centering
    \begin{subfigure}[b]{0.42\textwidth}
        \centering
        \includegraphics[width=\linewidth]{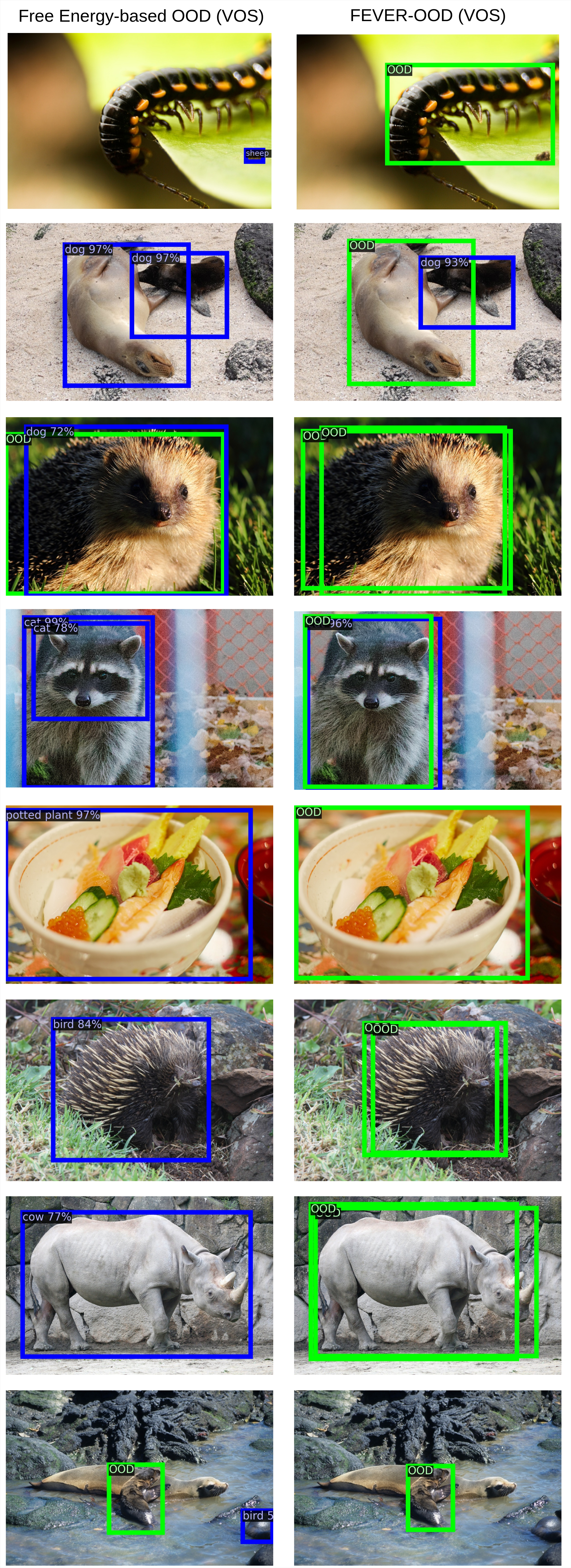}
        \caption{}
    \end{subfigure}%
    ~ 
    \begin{subfigure}[b]{0.42\textwidth}
        \centering
        \includegraphics[width=\linewidth]{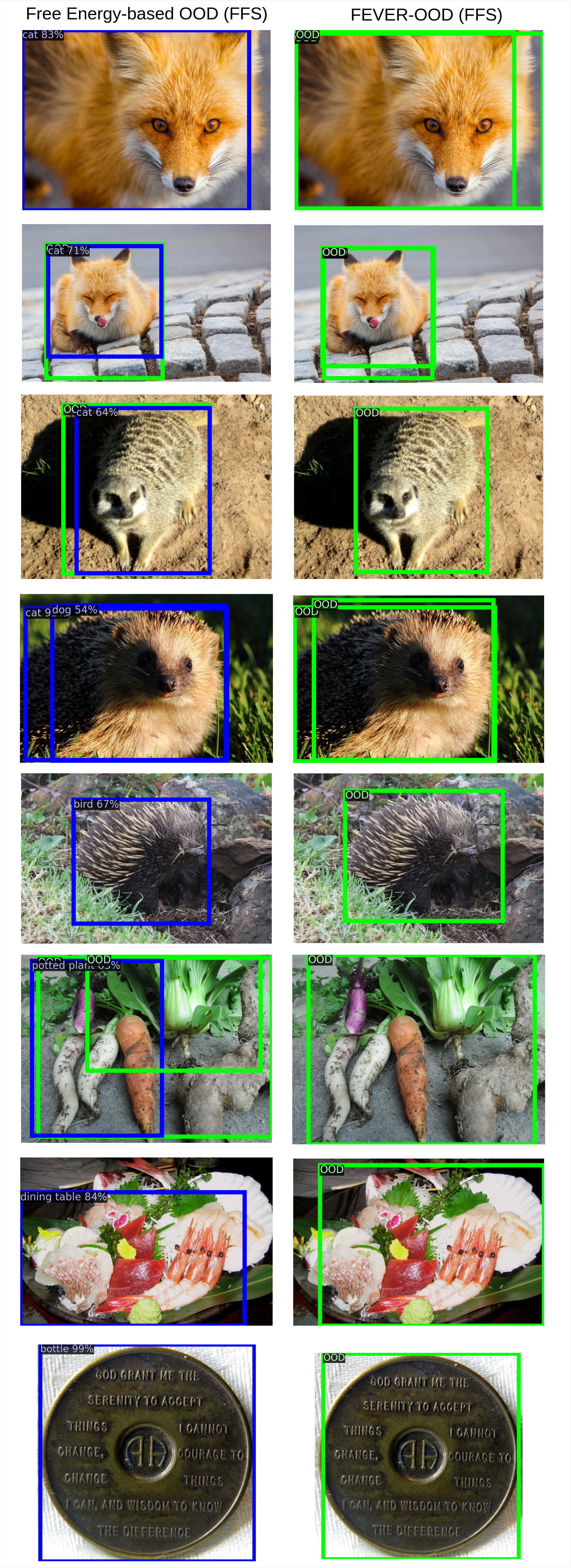}
        \caption{}
    \end{subfigure}
    \caption{Additional visualization of detected objects on the OOD images (from OpenImages \cite{kuznetsova2020open}) by free energy-based OOD (VOS) \cite{duvos}, free energy-based OOD (FFS) \cite{kumar2023normalizing} and FEVER-OOD (our approach). The in-distribution is PASCAL VOC \cite{voc_dataset} dataset. \textcolor{blue}{Blue}: OOD objects detected and mis-classified as being in-distribution. \textcolor{green}{Green}: the same OOD objects correctly detected as OOD by FEVER-OOD (ours).}
    \label{fig:open-images-more-qualitative}
\end{figure*}

\begin{figure*}[t!]
    \centering
    \begin{subfigure}[b]{0.42\textwidth}
        \centering
        \includegraphics[width=\linewidth]{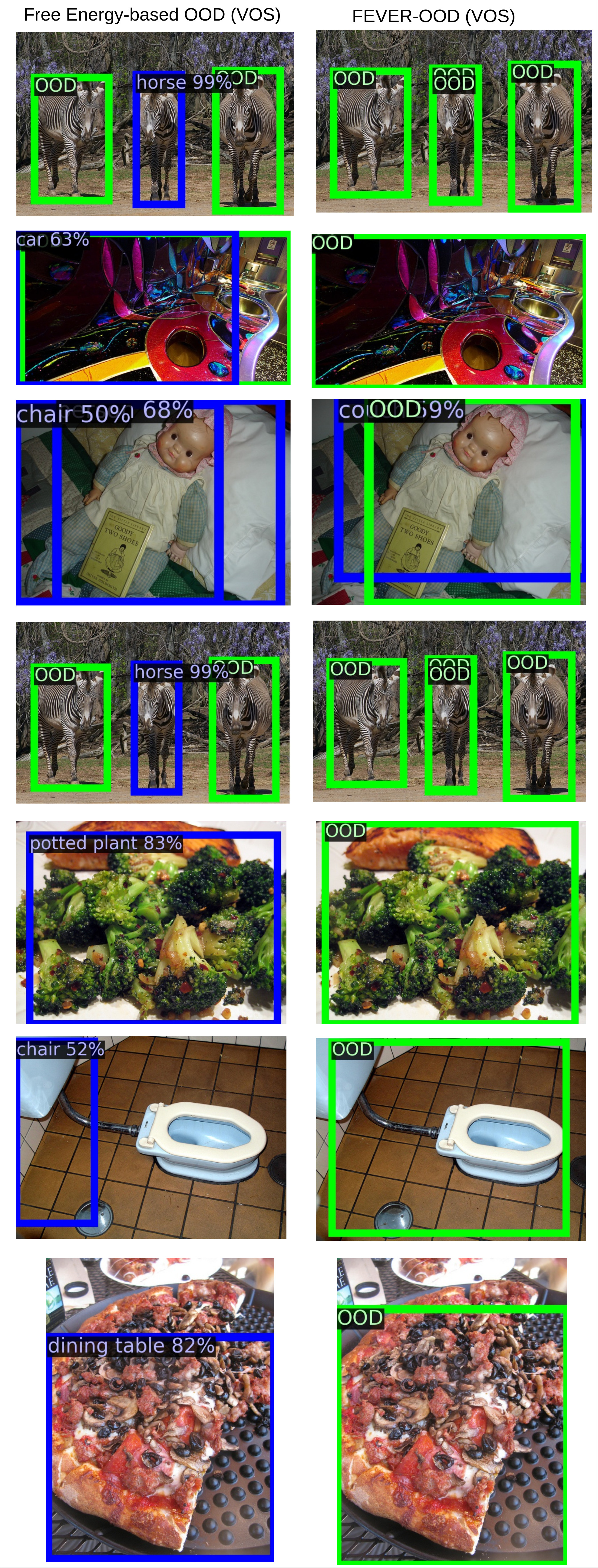}
        \caption{}
    \end{subfigure}%
    ~ 
    \begin{subfigure}[b]{0.42\textwidth}
        \centering
        \includegraphics[width=\linewidth]{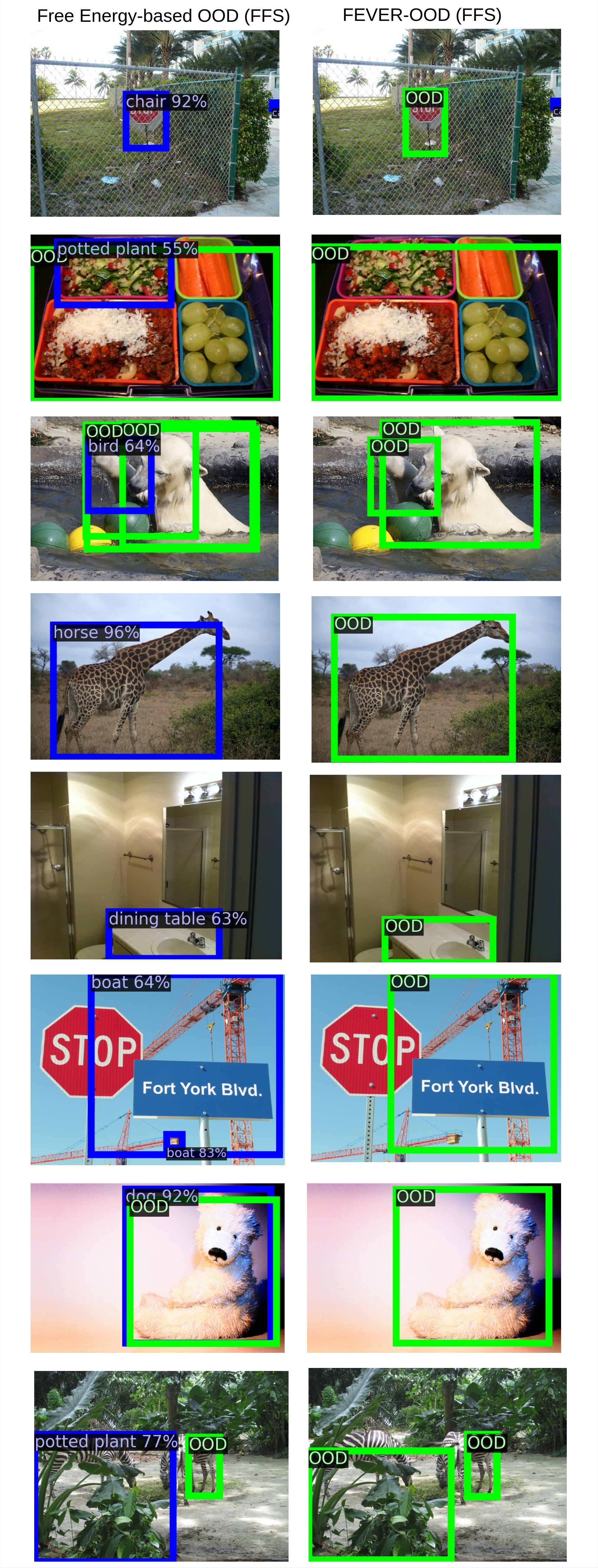}
        \caption{}
    \end{subfigure}
    \caption{Additional visualization of detected objects on the OOD images (from MS-COCO \cite{coco_dataset}) by free energy-based OOD (VOS) \cite{duvos}, free energy-based OOD (FFS) \cite{kumar2023normalizing} and FEVER-OOD (our approach). The in-distribution is PASCAL VOC \cite{voc_dataset} dataset. \textcolor{blue}{Blue}: OOD objects detected and mis-classified as being in-distribution. \textcolor{green}{Green}: the same OOD objects correctly detected as OOD by FEVER-OOD (ours).}
    \label{fig:ms-coco-more-qualitative}
\end{figure*}

Additional qualitative examples for object-level OOD detection using VOS \cite{duvos} and FFS \cite{kumar2023normalizing} models trained with and without FEVER-OOD with PASCAL VOC as in-distribution are shown in \cref{fig:open-images-more-qualitative} for OpenImages \cite{kuznetsova2020open} as OOD, and in \cref{fig:ms-coco-more-qualitative} fos MS-COCO \cite{coco_dataset} as OOD. 


\section{Limitations and Potential Negative Impact}
Finally, this section discusses some limitations and potential negative impact of FEVER-OOD. We identify the following limitations:
\begin{itemize}
    \item FEVER-OOD does not entirely avoid the null space vulnerabilities. While we reduce the size of it, there might be some anomalies with large components in the feature space. 
    \item Careful fine tuning is needed in some instances, specially when reducing the null space significantly. We did not identify any condition to estimate the regularizer weight \textit{a priori}.
    \item  While our analysis show a large change in Energy for far anomalies (OOD samples), we did not test the performance of FEVER-OOD in this cases.
\end{itemize}
Finally, our work might have some potential negative impact. For instance, the exploration of these vulnerabilities might allow for tailored generated outliers that fool models based on free energy OOD detection. Additionally, we trained several models with different in-distribution datasets to perform the ablation studies, having a negative environmental effect due to the large power consumption for GPU training.



\clearpage

\clearpage

\end{document}